\documentclass[lettersize,journal]{IEEEtran}
\usepackage{amsmath,amsfonts,amssymb}
\usepackage{booktabs}
\usepackage{multirow}
\usepackage{array}
\usepackage{tabularx}
\usepackage{makecell}
\usepackage{siunitx}
\usepackage{graphicx}
\usepackage{subcaption}   
\usepackage{wrapfig}
\usepackage{adjustbox}
\usepackage{tikz}
\usepackage{pgfplots}
\usepackage{algorithm}
\usepackage{algpseudocode}  
\usepackage{enumitem}
\usepackage{setspace}
\usepackage{verbatim}
\usepackage{textcomp}
\usepackage{url}
\usepackage{xcolor}
\usepackage[table]{xcolor}
\usepackage{stfloats}
\usepackage{listings}
\usepackage[most]{tcolorbox}
\usepackage{cite}
\definecolor{eclipseGreen}{RGB}{63, 127, 95}  
\algrenewcommand{\algorithmiccomment}[1]{\hfill \textcolor{eclipseGreen}{\footnotesize // #1}}

\def\BibTeX{{\rm B\kern-.05em{\sc i\kern-.025em b}\kern-.08em
    T\kern-.1667em\lower.7ex\hbox{E}\kern-.125emX}}

\newcommand{\tool}{\textsc{BayesWarp}}

\begin{document}

\title{Testing Neural Networks via Bayesian-Guided Exploration of Decision Landscapes}
\author{
  Bin Duan$^{1}$,
  Meiru Che$^{2}$,
  Guowei Yang$^{1*}$ \\
  $^{1}$School of Electrical Engineering and Computer Science, The University of Queensland, Australia  \\
  $^{2}$College of Information and Communications Technology, Central Queensland University, Australia\\
  \{b.duan, guowei.yang\}@uq.edu.au, {m.che@cqu.edu.au}
}

\markboth{Transactions on Software Engineering}%
{Testing Neural Networks via Bayesian-Guided Exploration of Decision Landscapes}

\maketitle

\begin{abstract}

As neural networks are increasingly deployed in safety-critical domains, testing is essential to evaluate and improve their reliability. Existing testing methods, whether black-box or white-box, primarily use global mutation or coverage-guided strategies, both of which struggle to efficiently uncover diverse model failures while remaining proximate to the original data distribution and semantics. We propose \tool, a testing framework that addresses this limitation by mutating decision-critical input regions identified via interpretable saliency techniques and adaptively guiding the testing process using an uncertainty-aware Bayesian Optimization strategy, enabling the discovery of diverse failures while preserving distributional and semantic proximity to the original data. Evaluation on MNIST, CIFAR-10, and ImageNet across six neural network models shows that \tool\ improves failure discovery, failure diversity, test case quality, and critical neuron coverage under a fixed mutation budget. These results demonstrate that \tool\ improves testing effectiveness. Moreover, fine-tuning with the generated failure cases leads to improvements in model performance.

\end{abstract}


\section{Introduction}

\IEEEPARstart{D}{eep} Neural Networks (DNNs)~\cite{pinaya2020convolutional} are increasingly deployed in safety-critical applications such as autonomous driving~\cite{li2019stereo}, facial recognition~\cite{ben2021face}, and malware detection~\cite{ganesh2017cnn}, which makes testing essential for evaluating and improving their reliability.
To this end, DNN testing~\cite{pei2017deepxplore, odena2019tensorfuzz, lee2020effective, dola2024cit4dnn} has emerged as a method to assess the reliability of DNNs by generating test cases that expose their unexpected behaviors. Unlike traditional software, DNNs do not crash on invalid inputs and will still process and classify them, as long as the input format is correct.
Consequently, an effective DNN testing method must be viewed as a problem of efficiently exploring decision instability under limited mutation budgets, rather than merely generating misclassified inputs.
This requires systematically probing alternative decision behaviors while maintaining proximity to the original data distribution and semantics, so that revealed failures reflect intrinsic model weaknesses rather than arbitrary perturbations.
Distinct from adversarial attacks, which optimize misclassification under explicit threat-model or norm constraints, DNN testing focuses on systematically exploring a model’s decision behaviors to reveal diverse failures and internal states. Accordingly, existing DNN testing methods~\cite{lee2020effective, wang2022bet, dola2024cit4dnn, huang2024neuron} are evaluated in terms of failure quantity and diversity, test case quality, and coverage of internal behaviors rather than attack success rate. Formally, we view DNN testing as the problem of exploring a model’s decision landscape, defined as the mapping from input space to class confidence distributions. A critical challenge is that failures are not uniformly distributed in the input space, but are concentrated in regions where small perturbations can induce significant changes in decision boundaries, which we refer to as decision-critical regions. Therefore, effective testing should not aim to maximize global variation or coverage, but to efficiently explore these decision-critical regions to uncover diverse instability patterns under a limited query budget.

DNN testing approaches are categorized into black-box and white-box methods, depending on their access to internal information.
Black-box methods~\cite{odena2019tensorfuzz, wang2022bet, xie2019diffchaser, hu2023atom, dola2024cit4dnn} rely on model queries and observed outputs. Without gradients or internal states, they typically use random or search-based mutations, which lack guidance toward decision-critical regions and thus require many queries to find effective failures.
White-box methods~\cite{lee2020effective, pei2017deepxplore, guo2018dlfuzz, wang2021robot, guo2025white} exploit internal signals such as neuron activations, with a primary goal of improving neuron coverage. However, coverage maximization alone~\cite{lee2020effective, guo2025white} does not distinguish between decision-critical and irrelevant activations~\cite{harel2020neuron, yang2022revisiting}, which result in test cases that increase coverage but deviate from original data distributions and semantics.
Across both black-box and white-box settings, existing DNN testing methods largely rely on global mutations or coverage-driven strategies that optimize coverage or global variation rather than decision-level instability, wasting testing budget on decision-irrelevant regions and leading to redundant failures and inefficient exploration of model weaknesses.
A key practical challenge is that DNN testing must operate under a limited mutation budget while still exploring multiple competing failure directions. In this setting, purely random mutation fails to exploit historical observations, whereas single-direction optimization tends to over-commit to a locally dominant direction and repeatedly expose similar failures. What is needed is a testing strategy that can both exploit promising mutation trends and preserve uncertainty-aware exploration of alternative decision paths.

To address these limitations, we propose \tool, a white-box testing framework for decision-critical exploration. Rather than mutating the entire input space or pursuing coverage signals alone, \tool\ focuses testing on regions that are more likely to influence model decisions. Our design combines three complementary elements. First, we construct structured decision-critical regions from saliency maps through connected components, bounding-box merging, and region constraints, thereby transforming pixel-level attribution into a tractable mutation space. Second, we define a diversity-oriented testing objective with adaptive weighting and target-class scheduling to encourage exploration of multiple competing failure directions rather than a single dominant one. Third, we use a grid-parameterized Sparse Variational Gaussian Process (SVGP) surrogate within Bayesian optimization to support uncertainty-aware search under a fixed mutation budget. Together, these components enable localized, diversity-oriented, and budget-aware testing of neural networks.

We evaluate the effectiveness of \tool\ across three benchmark datasets (MNIST, CIFAR-10, and ImageNet) and six representative neural network models. Experimental results demonstrate that, compared to state-of-the-art white-box testing techniques, under fixed mutation budget, \tool\ efficiently uncovers a larger and more diverse set of failure-inducing test cases, achieves a higher failure-inducing seed rate. Moreover, the generated test cases remain close to the original data distribution while preserving high semantic similarity to their corresponding input seeds. 
While our approach is not designed to maximize conventional coverage metrics, it also attains strong critical-neuron-related coverage in the evaluated settings.
In summary, our contributions are as follows:

\noindent \textbf{Approach}. We propose \tool, a white-box testing method, introducing: \textbf{(i)} a decision-critical region localization mechanism that combines saliency maps, connected-component merging, and region constraints; \textbf{(ii)} a diversity-oriented testing objective with adaptive weighting and adaptive target-class scheduling; and \textbf{(iii)} a scalable mutation framework based on grid-parameterized SVGP Bayesian optimization.

\noindent \textbf{Tool}. We implement \tool\ as an automated testing framework for neural network models. We make \tool\ publicly available to facilitate reproducibility~\cite{Our}.

\noindent \textbf{Evaluation}. We evaluate on three benchmark datasets and six widely used models. Results show that \tool\ efficiently uncovers a broader spectrum of failure modes, achieves higher neuron coverage, and generates test cases that are distributionally close to the original dataset and semantically similar to their corresponding input seeds, leading to more effective performance improvements during fine-tuning than state-of-the-art testing techniques.

\section{Background}

\subsection{Localization of Decision-Critical Regions}

Various techniques have been developed to localize critical regions in an image that influence a model’s prediction, thereby enhancing interpretability by highlighting features relevant to classification decisions.
Representative methods include the following:
LRP~\cite{bach2015pixel} propagates relevance scores backward through network layers in a conservative manner, ensuring that total relevance is preserved.
Grad-CAM~\cite{selvaraju2017grad} computes gradients of the target class with respect to convolutional feature maps and averages them to produce heatmaps that localize decision-critical regions.
DeepLIFT~\cite{shrikumar2017learning} assigns contribution scores by comparing neuron activations to predefined reference activations and propagating these differences through the network.
Integrated Gradients~\cite{kapishnikov2021guided} quantifies pixel contributions by integrating gradients along a path interpolated between the input and a baseline.
SmoothGrad~\cite{smilkov2017smoothgrad} improves the stability of gradient-based attributions by averaging saliency maps obtained from noisy versions of the input.
Grad-CAM++~\cite{chattopadhay2018grad} extends Grad-CAM by incorporating higher-order derivatives to generate more fine-grained and class-discriminative saliency maps.
These techniques provide practical tools for identifying decision-critical regions used in our method.

\subsection{Bayesian Optimization}

Bayesian optimization~\cite{frazier2018tutorial} is an adaptive optimization strategy that employs Gaussian Processes (GPs)~\cite{deringer2021gaussian} to model the objective function in a non-parametric manner, providing probabilistic estimates and balancing exploration and exploitation to efficiently identify near-optimal solutions with few queries.
It relies on acquisition functions, such as Expected Improvement (EI)~\cite{zhan2020expected} and Upper Confidence Bound (UCB)~\cite{kaufmann2012bayesian}, to select informative sampling points and iteratively refine the search.
However, directly applying Bayesian optimization to DNN testing is challenging due to the high dimensionality of input spaces and the complex, non-smooth decision landscapes of neural networks~\cite{ryan2016review}, which can lead to noisy exploration and inefficient sampling. This challenge is particularly relevant to testing, where each model query consumes budget and where the search objective may be locally misleading or multi-modal. In such settings, uncertainty-aware search is preferable to purely greedy optimization because it can avoid over-committing to a single mutation direction. To mitigate these issues, we adopt an SVGP surrogate~\cite{titsias2009variational} and restrict the mutation space to decision-critical regions, which reduces the effective search dimensionality and biases the optimization toward regions that are more likely to induce failures.

\begin{figure*}[t!]
\centering
\includegraphics[width=1\linewidth]{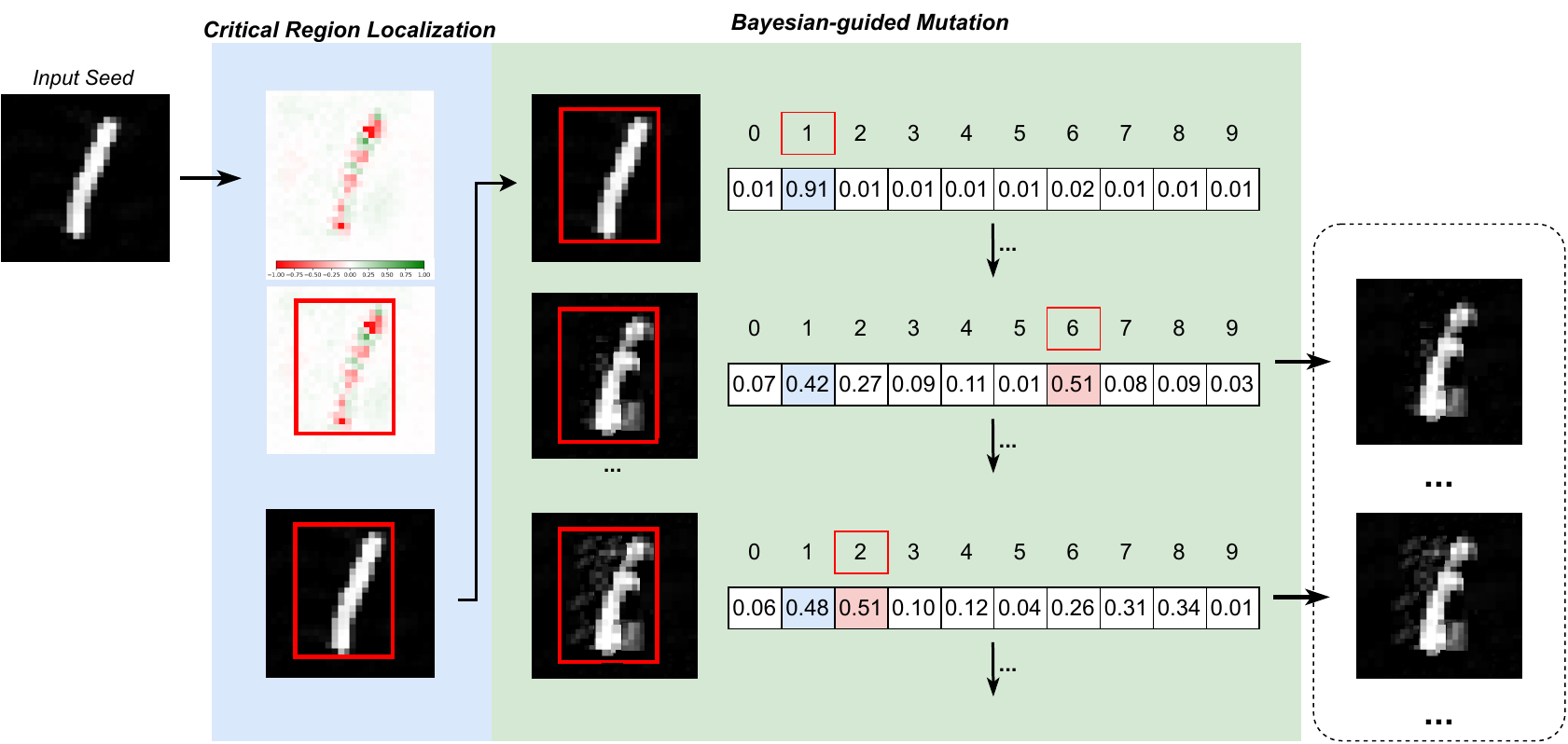}
\vspace{-1mm}
\caption{{Overview of \tool}}
\label{fig:overview}
\vspace{-5mm}
\end{figure*}

\section{Method}

\tool\ proposes a Bayesian-guided white-box testing framework that explores instabilities in a model’s decision landscape. 
The core idea is to (1) localize mutation to decision-critical input regions identified by saliency-based interpretability, and (2) adaptively guide mutations within this localized space via Bayesian optimization to expose diverse failure modes. 
To make the search efficient under a fixed mutation budget, \tool\ parameterizes localized mutations on a coarse grid and accelerates Bayesian optimization with an SVGP surrogate.
Fig~\ref{fig:overview} shows the overview of \tool. {Given an input seed, the framework first localizes decision-critical regions via saliency analysis and then performs Bayesian-guided mutations within these regions to efficiently generate diverse failure-inducing test cases by exploring different target classes.}
Sec~\ref{sec:locate} describes the process for locating and refining decision-critical regions using interpretability methods;
Sec~\ref{sec:bayesian} describes the formal construction of the objective function, the adaptive class selection, and a surrogate-based Bayesian optimization procedure based on SVGP;
Sec~\ref{sec:code} presents the algorithm of \tool.

\subsection{Critical Region Localization}
\label{sec:locate}

We identify and localize decision-critical regions that are highly influential to model decisions.
A saliency map $H(x) \in \mathbb{R}^{H \times W}$ is computed using a saliency-based interpretability method and normalized to $[0,1]$.
We retain the top-$\alpha$ proportion of salient pixels by thresholding $H(x)$ at the $(1-\alpha)$ quantile:
\[
T_{\alpha} = \mathrm{Quantile}(H(x), 1-\alpha), \quad
M = \mathbf{1}(H(x) > T_{\alpha}),
\]
where $\alpha$ explicitly controls the proportion of preserved salient pixels.
From the binary mask $M$, we extract 8-connected components, including diagonal adjacency, and discard components with region smaller than $A_{\min}$ to remove spurious saliency fragments.
Each remaining component is enclosed by a minimal bounding box.  
To reduce fragmentation, two boxes are merged whenever they substantially overlap or lie sufficiently close:
\[
\mathrm{IoU}(B_i, B_j) > \tau_{\text{iou}}
\quad \text{OR} \quad
d_{\mathrm{centroid}}(B_i, B_j) < d_{\max},
\]
where $\tau_{\text{iou}} \in (0,1)$ and $d_{\max} > 0$ control overlap and spatial proximity.  
These two criteria reflect complementary notions of locality: geometric intersection and saliency continuity across neighboring salient regions.
Let $\mathcal{B} = \{B_1,\ldots,B_K\}$ denote the merged boxes.  
To avoid over-merging that would undermine localization specificity, we enforce a coverage constraint on the mutation region.
Specifically, the merged boxes are ranked by their saliency mass, defined as the sum of saliency values within each box, and selected in a greedy manner until the total covered area reaches a predefined budget:
\[
\frac{\mathrm{Region}\!\left(\bigcup_{B_i \in \mathcal{B}'} B_i\right)}
{\mathrm{Region}(\text{image})} \le \rho,
\]
where $\mathcal{B}' \subseteq \mathcal{B}$ denotes the selected subset and $\rho \in (0,1)$ bounds the proportion of input space allowed for mutation.
The union of the selected boxes forms the final critical region $R = \bigcup_{B_i \in \mathcal{B}'} B_i$.
All subsequent Bayesian optimization steps restrict mutations to $R$, reducing the search space while preserving focus on the decision-critical input regions. Based on these steps, \tool\ transforms saliency into merge-aware region constraints, producing mutation spaces that are both decision-relevant and amenable to probabilistic optimization.

\subsection{Bayesian-guided Mutation}
\label{sec:bayesian}

After locating the decision-critical region of input seeds, we perform uncertainty-aware testing using a Bayesian-guided mutation strategy to adaptively guide mutation directions. The goal is to explore different prediction behaviors of the model, focusing on uncovering a diverse spectrum of failure modes while encouraging the mutated inputs to remain close to the original data distribution and semantics.
To achieve this, Bayesian optimization iteratively proposes mutation parameters by maximizing an acquisition function. The underlying objective value $f(x(u))$ is modeled by an SVGP surrogate defined over the low-dimensional mutation parameters $u$. This probabilistic approach balances exploration and exploitation through the acquisition function, where exploration searches for new failure-inducing mutation directions and exploitation refines mutations that have already revealed diverse erroneous model behaviors. This process is guided by a composite objective function that adaptively balances model confidence changes across classes, thereby encouraging the exploration of diverse failure modes. Unlike purely greedy search, which tends to follow a single locally dominant direction, Bayesian optimization retains an explicit exploration mechanism through predictive uncertainty, making it better suited to uncovering multiple competing failure directions. In this context, Bayesian optimization guide testing toward diverse decision instabilities.

\subsubsection{Define the Objective Function}

To explore diverse prediction behaviors, we define an objective that adaptively trades off suppressing the original class confidence and promoting a selected target class confidence.
 The objective function is formulated as:

\begin{equation}
f(x) = \lambda(x) \cdot \textit{Conf}_{\textit{tg}}(x) - (1 - \lambda(x)) \cdot\textit{Conf}_{\textit{og}}(x),
\end{equation}
where $\textit{Conf}_{\textit{tg}}(x)$ and $\textit{Conf}_{\textit{og}}(x)$ denote the softmax probabilities assigned by the model to the target class and the original class, respectively. The adaptive weight $\lambda(x)$ balances the two terms based on their relative progression during mutation:
\begin{equation}
\lambda(x)=\frac{\textit{Conf}_{tg}(x)}{\textit{Conf}_{tg}(x)+\textit{Conf}_{og}(x)}.
\end{equation}
This normalization provides a smooth transition between the two objectives.
At early stages, the target class confidence is much lower than that of the original class, resulting in a small $\lambda(x)$ and placing more emphasis on suppressing $\textit{Conf}_{og}(x)$ to destabilize the original decision and explore alternative decision directions.
As mutations progress and the target class confidence increases relative to the original class, $\lambda(x)$ grows accordingly, gradually shifting the objective toward promoting $\textit{Conf}_{tg}(x)$ to induce prediction divergence.
This adaptive weighting avoids abrupt objective changes and enables a smooth search trajectory across target classes, from weakening the dominant decision to transitioning toward alternative classes.

\subsubsection{Adaptive Target Class Selection}
To expose diverse failure modes, we iteratively select the target class $tg_k$ with the highest initial confidence among unexplored alternatives:
\begin{equation}
tg_k = \arg\max_{tg \ne og,\; tg \notin \mathcal{U}_{k-1}} \textit{Conf}_{tg}(x_0)
\end{equation}
where $x_0$ is the input seed, $og$ is the original prediction of $x_0$, and $\mathcal{U}_{k-1}$ is the set of previously selected target classes.
For each selected $tg_k$, we perform mutation under a fixed per-class mutation budget.
Notably, the objective continues guiding exploration within the same target class even after failures are induced, enabling the discovery of multiple failure-inducing test cases for the same $tg_k$ rather than terminating after a single prediction change.
We switch to the next unexplored target class when the allocated budget for the current class is exhausted. The next target class is selected as the unexplored class with the highest initial confidence.
By prioritizing target classes with higher initial confidence and allocating a fixed budget to each selected class, this strategy reduces over-focusing on a single class while avoiding premature termination after isolated failures.
In practice, it encourages broader coverage of candidate classes and helps produce a more diverse set of failure-inducing test cases.

\subsubsection{SVGP Approximation}

Having defined the exploration objective and the adaptive target class selection strategy, the remaining challenge is to efficiently search the mutation space to optimize this objective under a limited testing budget. 
Standard Gaussian Process (GP)-based Bayesian Optimization can become computationally prohibitive as the number of evaluated mutations grows.
To address this, we employ a SVGP surrogate~\cite{titsias2009variational}, which avoids the complexity of full GPs and enables scalable optimization in high-dimensional testing scenarios.
We define a set of inducing points $Z = \{z_i\}_{i=1}^{m}$ in the low-dimensional mutation parameter space, where $m$ controls the approximation capacity of the SVGP surrogate.
Concretely, mutations are parameterized on a coarse $n \times n$ spatial grid covering the localized critical region.
Each grid cell corresponds to a low-dimensional mutation variable, and the resulting grid-level mutations are upsampled to pixel resolution via bilinear interpolation.
This parameterization significantly reduces the effective search dimension while maintaining spatial coherence of the mutations.
Since the decision-critical region $R$ is input-dependent, the mutation parameterization is constructed with respect to the current mutation trajectory. The inducing point set $Z$ is initialized accordingly for the corresponding mutation space. Bayesian optimization is performed in a trajectory-wise manner, during which $R$ and the associated mutation parameterization remain fixed, and the SVGP surrogate is incrementally updated using newly observed objective values.

For a candidate mutation parameter vector $u$, the objective function $f(x(u))$ is approximated using the SVGP surrogate, which provides a predictive mean $\mu(u)$ and uncertainty $\sigma(u)$.
These quantities are used by the acquisition function to guide mutation selection.
Specifically, we assume a variational distribution over the inducing variables, $q(\mathbf{v}) = \mathcal{N}(\mathbf{m}, \mathbf{S})$, following the standard SVGP formulation~\cite{titsias2009variational}.
The predictive mean and variance are given by:
\begin{equation}
\mu(u) = k(u,Z)^\top K_{ZZ}^{-1} \mathbf{m},
\end{equation}
\begin{equation}
\sigma^2(u) = k(u,u) + k(u,Z)^\top K_{ZZ}^{-1}
(\mathbf{S} - K_{ZZ}) K_{ZZ}^{-1} k(Z,u),
\end{equation}
where $k(\cdot,\cdot)$ denotes a standard positive-definite kernel function, $K_{ZZ}$ is the kernel matrix over inducing points, and $\mathbf{m}$ and $\mathbf{S}$ are the mean and covariance of the variational distribution $q(\mathbf{v})$, respectively.
In this formulation, $\mu(u)$ serves as a surrogate estimate of the objective value, while $\sigma(u)$ captures the model uncertainty around $u$.
These estimates are used to evaluate and rank candidate mutations via the acquisition function, rather than to perform exact posterior inference.

To balance exploration and exploitation during mutation search, we adopt an upper-confidence bound (UCB) acquisition function:
\begin{equation}
a(u) = \mu(u) + \kappa \, \sigma(u),
\end{equation}
where $\kappa$ controls the exploration--exploitation trade-off.
This formulation favors candidates with either high estimated objective values or high uncertainty.

The SVGP surrogate is trained by maximizing the standard evidence lower bound (ELBO)~\cite{titsias2009variational} on observed objective values, allowing efficient surrogate fitting under the fixed mutation budget.

\subsubsection{Mutation Constraints and Updates}

To ensure that the generated test cases remain close to the original data distribution, we apply element-wise clipping after each mutation step to keep pixel values within a bounded and controlled range:
\begin{equation}
P_{\min} - \eta(P_{\max}-P_{\min}) \leq P' \leq P_{\max} + \eta(P_{\max}-P_{\min}),
\end{equation}
where $P_{\min}$ and $P_{\max}$ denote the minimum and maximum pixel values, and $\eta$ is a small scaling factor. This symmetric relaxation permits limited exploration slightly beyond the original data range within a controlled bound.

Each mutation is parameterized by a low-dimensional mutation parameter vector $u$ defined on the grid-based representation of the decision-critical region. Here, $u$ represents additive pixel mutation amplitudes defined on a spatial grid, rather than parameters of global image transformations. We obtain the corresponding pixel-level mutation by applying a bilinear upsampling operator $\mathcal{I}(\cdot)$:
\(
\delta x = \mathcal{I}(u),
\)
where $\mathcal{I}(\cdot)$ upsamples the $n \times n$ grid to the input resolution while preserving spatial smoothness.
At each mutation step, the ideal update would be obtained by solving \(\delta u = \arg\max_{\delta}\, a(u + \delta) \), but this continuous optimization is computationally prohibitive. Thus, we approximate acquisition maximization by sampling a small candidate set $\Delta = \{\delta_j\}_{j=1}^{S}$ in the mutation space, where $S$ denotes the number of sampled candidate mutations, and selecting the top-ranked candidate according to the acquisition function:
\begin{equation}
\delta u = \arg\max_{\delta \in \Delta}\bigl[\mu(u_t + \delta) + \kappa\, \sigma(u_t + \delta)\bigr],
\end{equation}
where $\mu(\cdot)$ and $\sigma(\cdot)$ denote the predictive mean and standard deviation of the SVGP surrogate. The selected $\delta u$ is then mapped to a pixel-level update $\delta x$, and the input is updated as
\begin{equation}
x_{t+1} = x_t + \delta x,
\end{equation}
Correspondingly, $u_{t+1}=u_t+\delta u$, and the new observation $(u_{t+1}, f(x_{t+1}))$ is added to refine the surrogate model.
To prevent premature convergence to local optima, if the improvement $\Delta f = f(x_{t+1}) - f(x_t)$ falls below a small threshold $\epsilon$, we inject a small random mutation:
\begin{equation}
x_{t+1} =
\begin{cases}
    x_t + \delta x, & \text{if } \Delta f > \epsilon \\
    x_t + \delta x + \beta \cdot \mathcal{N}(0, I), & \text{otherwise}
\end{cases}
\end{equation}
where $\beta$ controls the noise magnitude and encourages broader exploration of the mutation space.


\subsection{Algorithm}
\label{sec:code}

\begin{algorithm}[t!]
\setstretch{0.9}
\footnotesize
\caption{\tool}
\label{alg:bayesian_mutation}
\begin{algorithmic}[1]
\State \textbf{Input:} Model $\mathcal{M}$, seed $x_0$, max classes $C$
\State \textbf{Output:} Mutated test cases $\mathcal{X}_{test}$
\State Initialize $\mathcal{X}_{test} \gets \emptyset$, $\mathcal{D} \gets \emptyset$
\State $og \gets \arg\max_{c\in\{1,\ldots,C\}} \textit{Conf}_{c}(x_0)$
\State Identify decision-critical region $R$ via interpretability methods (Sec.~\ref{sec:locate})
\State Construct grid parameterization on $R$ and initialize SVGP surrogate (Sec.~\ref{sec:bayesian})
\State $\mathcal{U} \gets \{1,\ldots,C\}\setminus\{og\}$
\While{$\mathcal{U} \neq \emptyset$}
    \State $tg \gets \arg\max_{c \in \mathcal{U}} \textit{Conf}_{c}(x_0)$
    \State $\mathcal{U} \gets \mathcal{U} \setminus \{tg\}$; \ $x \gets x_0$; \ $u \gets \mathbf{0}$
    \While{per-class mutation budget not exhausted}
        \State $\lambda(x) \gets \frac{\textit{Conf}_{tg}(x)}{\textit{Conf}_{tg}(x)+\textit{Conf}_{og}(x)}$
        \State $f(x) \gets \lambda(x)\cdot \textit{Conf}_{tg}(x) - (1-\lambda(x))\cdot \textit{Conf}_{og}(x)$
        \State Update $\mathcal{D} \gets \mathcal{D} \cup \{(u, f(x))\}$ and update SVGP surrogate
        \State Sample candidate steps $\Delta=\{\delta_j\}_{j=1}^{S}$; \
        $\delta u \gets \arg\max_{\delta\in\Delta}\big[\mu(u+\delta)+\kappa\,\sigma(u+\delta)\big]$
        \State $u' \gets u+\delta u$; \ $\delta x \gets \mathcal{I}(u')$; \ $x' \gets x + \delta x$
        \State Compute $f(x')$
        \If{$|f(x') - f(x)| \leq \epsilon$}
            \State $x' \gets x' + \beta \cdot \mathcal{N}(0, I)$
        \EndIf
        \If{$\arg\max_{c\in\{1,\ldots,C\}} \textit{Conf}_{c}(x') \neq og$}
            \State $\mathcal{X}_{test} \gets \mathcal{X}_{test} \cup \{x'\}$
        \EndIf
        \State $x \gets x'$; \ $u \gets u'$
    \EndWhile
\EndWhile
\State \textbf{Return} $\mathcal{X}_{test}$
\end{algorithmic}
\end{algorithm}

Algorithm~\ref{alg:bayesian_mutation} shows the key steps of \tool\ in generating failure-inducing test cases through Bayesian-guided mutation restricted to decision-critical input regions.

Starting from a seed input $x_0$, the algorithm determines the original prediction class $og$, initializes the test set, and identifies the decision-critical region $R$ using interpretability methods, before constructing a grid-based parameterization on $R$ and initializing an SVGP surrogate for Bayesian optimization (lines~3--6).
The algorithm maintains a set of unexplored alternative classes and iteratively selects a target class $tg$ with the highest confidence among them. For each selected class, the mutation process is reset from the original seed with zero-initialized mutation parameters (lines~7--10).
Within the per-class mutation budget, a composite objective function $f(x)$ is evaluated to adaptively suppress the original class while promoting the target class via a dynamic weight $\lambda(x)$. The current observation is used to update the SVGP surrogate, and Bayesian optimization proposes the next mutation step by maximizing a UCB-based acquisition function over sampled candidates (lines~11--15).
The selected mutation is mapped to the input space to generate a new input $x'$, whose objective value is evaluated. If the objective improvement is insufficient, a small stochastic noise is injected to encourage exploration (lines~16--19). Whenever $x'$ induces a prediction different from the original class, it is recorded as a failure-inducing test case, while the search continues within the same target class (lines~20--23).
After exhausting the mutation budget for the current target class, the algorithm proceeds to the next unexplored class and repeats the above process (lines~24--27).

\section{Evaluation}
\subsection{Research Questions}
Our evaluation aims to answer the following research questions:
\begin{itemize}
  \item[\textbf{RQ1:}] How effective is \tool\ in testing neural networks?
  \item[\textbf{RQ2:}] Can the generated test cases be leveraged to improve model performance?
  \item[\textbf{RQ3:}] What is the impact of each main component in \tool?
\end{itemize}

For \textbf{RQ1}, we evaluate the effectiveness of \tool\ in generating test cases from multiple complementary perspectives, including failure discovery capability and efficiency, failure diversity, test case quality, and the exploration of internal neurons in a white-box setting. Together, these aspects provide a comprehensive assessment of \tool’s testing effectiveness.
For \textbf{RQ2}, we further investigate the practical value of the generated test cases by examining whether they can improve model performance through fine-tuning.
For \textbf{RQ3}, we conduct an ablation study to analyze the impact of the main components of \tool, focusing on decision-critical region localization and Bayesian-guided search.

\begin{table}
\centering
\footnotesize
\setlength{\tabcolsep}{26pt}
\renewcommand{\arraystretch}{0.5}
\vspace{-1mm}
\caption{Target Datasets and Models}
\vspace{-1mm}
\label{tab:models}
\begin{tabular}{lll}
\toprule
\textbf{Dataset}   & \textbf{Model} & \textbf{Accuracy} \\
\midrule
\multirow{2}{*}{MNIST}     & LeNet-4   & 97.50\% \\
                           & LeNet-5   & 98.12\% \\
\midrule
\multirow{2}{*}{CIFAR-10}  & VGG16     & 89.43\% \\
                           & ResNet18  & 88.31\% \\
\midrule
\multirow{2}{*}{ImageNet}  & VGG19     & 74.22\% \\
                           & ResNet50  & 75.52\% \\
\bottomrule
\end{tabular}
\vspace{-3mm}
\end{table}

\subsection{Experimental Setup}
\label{experimental}
\noindent\textbf{Target Models and Datasets.} 
Following~\cite{xie2019diffchaser, wang2022bet, lee2020effective, guo2018dlfuzz}, we selected three of the most widely used datasets and six representative neural network architectures as our experimental subjects. As shown in Table~\ref{tab:models}, MNIST~\cite{lecun2010mnist} consists of $28\times28$ grayscale images with 60,000 training and 10,000 test samples. 
We trained LeNet-4 and LeNet-5~\cite{lecun1998gradient} from scratch on MNIST for 20 epochs using the Adam optimizer with a learning rate of $1\times10^{-3}$.
CIFAR-10~\cite{krizhevsky2009learning} contains $32\times32$ color images from 10 classes, with 50,000 training and 10,000 test samples. 
We adopted ImageNet-pretrained VGG16~\cite{simonyan2014very} and ResNet18~\cite{he2016deep}, and fine-tuned all layers on CIFAR-10 for 30 epochs using Adam with a learning rate of $1\times10^{-4}$.
ImageNet~\cite{deng2009imagenet} contains $224\times224$ RGB images across 1,000 object classes, with approximately 1.2 million training samples and 50,000 validation samples.
We select ImageNet-pretrained VGG19~\cite{simonyan2014very} and ResNet50~\cite{he2016deep} as target models for evaluation. Unless otherwise specified, all training-based evaluations (e.g., fine-tuning) use the Adam optimizer with a learning rate of $1\times10^{-4}$.
For each dataset, we randomly selected 100 correctly classified training samples as input seeds, and evaluated all methods on the same seed set.

\noindent\textbf{Variants of \tool.} 
To evaluate the effectiveness of \tool, we implemented \tool\ based on three representative saliency-based interpretability techniques, Grad-CAM~\cite{selvaraju2017grad}, Integrated Gradients~\cite{kapishnikov2021guided}, and SmoothGrad~\cite{smilkov2017smoothgrad} to identify decision-critical regions for guided mutation: \tool-C (Grad-CAM), \tool-I (Integrated Gradients), and \tool-S (SmoothGrad).

\noindent\textbf{Environment.}
All experiments are performed on a Ubuntu 24.04.3 LTS system with an AMD Ryzen Threadripper PRO 7985WX (64-core) CPU, and NVIDIA RTX 6000 Ada Generation GPU.

\noindent\textbf{Hyperparameters.}
We retain the $\alpha = 0.1$, merge boxes with $\tau_{\text{iou}}=0.3$ and $d_{\max}=6$, and cap the merged region at $\rho=0.6$ of the image.
For Bayesian-guided mutation, we use $S=32$ per iteration, apply pixel-range relaxation with $\eta=0.1$, inject random noise with $\beta\in[0.01,0.05]$ when the improvement falls below a dataset-specific threshold $\epsilon$, and set the exploration weight $\kappa = 1$.
At each iteration, candidate updates $\delta$ are sampled independently from the same bounded uniform distribution in the grid parameter space, with accumulated grid parameters clipped to a fixed range.
The SVGP surrogate is updated once per iteration by maximizing the ELBO using Adam.
For MNIST, we set $A_{\min}=25$, $\epsilon=5\times10^{-4}$, the grid parameter $n = 1$, and use $m=64$ inducing points in the SVGP surrogate.
For CIFAR-10, we set $A_{\min}=50$, $\epsilon=10^{-3}$, $n = 2$, and $m=128$.
For ImageNet, we set $A_{\min}=250$, $\epsilon=5\times10^{-3}$, $n = 4$, and $m=256$.

\noindent\textbf{Baselines.}
We compare \tool\ with three representative white-box DNN testing methods with publicly available implementations: ADAPT~\cite{lee2020effective}, NSGen~\cite{huang2024neuron}, and SUNTest~\cite{guo2025white}. We selected these baselines because they are among the most relevant prior white-box methods and can be reproduced reliably on our infrastructure.
To improve comparability, we control the mutation budget, seed set, and prediction-inconsistency oracle across methods whenever the target setting can be executed reliably. However, some prior white-box methods rely on architecture-dependent neuron-selection rules, thresholding mechanisms, or internal heuristics that are tightly coupled with their original dataset--model settings. Directly porting such methods across datasets or architectures may alter their intended behavior and introduce additional confounding factors.
Therefore, we distinguish between fully controlled comparisons and reference comparisons. When a baseline can be executed reliably under a target setting, we compare it under the same budget, seed set, and oracle. When direct transfer would require modifying dataset-specific or architecture-specific mechanisms, we instead evaluate the method under the dataset--model settings reported in its original paper and interpret the comparison as a reference comparison rather than a perfectly uniform replication. Concretely, ADAPT is evaluated on MNIST and ImageNet, SUNTest on MNIST and CIFAR-10, and NSGen on CIFAR-10 and ImageNet.

\noindent\textbf{Mutation Budget.} 
Following prior work~\cite{huang2024neuron}, all methods are evaluated under a fixed budget of 10,000 mutations per input seed, using identical seeds and the same prediction-inconsistency oracle, to ensure a fair comparison in terms of query budget.

\subsection{Metrics}
\label{metrics}

The generated test cases are evaluated by feeding them into the target DNN models.
We adopt a prediction-inconsistency oracle, where a test case is considered to reveal a failure if it induces a different prediction from that of the original seed. Following prior DNN testing studies~\cite{lee2020effective, wang2022bet, wang2021robot, guo2025white}, such failures are defined as decision-stability violations rather than semantic correctness violations, reflecting excessive sensitivity in the model’s decision behavior, which is a core concern in neural network testing.
Following previous works~\cite{wang2022bet, xie2019diffchaser, lee2020effective, dola2024cit4dnn, huang2024neuron}, we use the following metrics to evaluate test cases. 

\noindent\textbf{\textit {Number of Failures (NoF)}}, as used in previous works~\cite{wang2022bet, xie2019diffchaser, lee2020effective, dola2024cit4dnn}, measures the total number of generated test cases identified as failure events by the inconsistency-based oracle, corresponding to prediction changes with respect to the original seeds.

\noindent\textbf{\textit {Failure-inducing Seed Rate (FSR)}}, as used in previous works~\cite{wang2022bet, lee2020effective}, reports the fraction of input seeds for which at least one failure test case is successfully generated. 

\noindent\textbf{\textit{Time per Failure (TPF)}} measures practical failure discovery efficiency, defined as the total wall-clock testing time divided by the number of failure events across all seeds, with all methods executed on the same environment.

\noindent\textbf{\textit {Diversity of Failures (DoF)}}, as used in previous works~\cite{wang2022bet, xie2019diffchaser, lee2020effective, dola2024cit4dnn},  measures the number of distinct predicted classes different from the seed prediction observed among all failure-inducing test cases. 

\noindent\textbf{\textit{Fréchet Inception Distance (FID)}}, as used in prior work~\cite{dola2024cit4dnn}, quantifies distributional proximity between generated test cases and the original input data. FID is computed at the dataset level using features extracted from a pretrained Inception-V3 network.

\begin{table*}[t!]
\centering
\footnotesize
\caption{Comparison Results on NoF$\uparrow$, FSR$\uparrow$, and TPF$\downarrow$.}
\vspace{-1mm}
\label{table:NoF-TPF-FSR}
\setlength{\tabcolsep}{7pt} 
\renewcommand{\arraystretch}{0.5}
\begin{tabular}{l | ccc |ccc |ccc |ccc |ccc}
\toprule
& \multicolumn{15}{c}{\textbf{MNIST}} \\
\cmidrule(lr){2-16}
& \multicolumn{3}{c}{ADAPT}
& \multicolumn{3}{c}{SUNTest}
& \multicolumn{3}{c}{\tool-C}
& \multicolumn{3}{c}{\tool-I}
& \multicolumn{3}{c}{\tool-S} \\
\cmidrule(lr){2-16}
\textbf{Model}
& NoF & FSR & TPF
& NoF & FSR & TPF
& NoF & FSR & TPF
& NoF & FSR & TPF
& NoF & FSR & TPF \\
\midrule
LeNet-4
& 1818.8 & 0.21 & 0.52
& 1285.1 & 0.15 & 1.10
& \cellcolor{gray!40}2362.0 & \cellcolor{gray!40}0.38 & \cellcolor{gray!40}0.43
& \cellcolor{gray!40}2624.5 & \cellcolor{gray!40}0.36 & \cellcolor{gray!40}0.39
& \cellcolor{gray!40}\textbf{3190.4}
& \cellcolor{gray!40}\textbf{0.47}
& \cellcolor{gray!40}\textbf{0.29} \\

LeNet-5
& 1544.1 & 0.24 & 0.39
& 1648.3 & 0.18 & 0.86
& \cellcolor{gray!40}2552.1 & \cellcolor{gray!40}0.34 & 0.39
& \cellcolor{gray!40}3656.0 & \cellcolor{gray!40}0.33 & \cellcolor{gray!40}0.27
& \cellcolor{gray!40}\textbf{3828.9}
& \cellcolor{gray!40}\textbf{0.49}
& \cellcolor{gray!40}\textbf{0.25} \\

\midrule\midrule
& \multicolumn{15}{c}{\textbf{CIFAR-10}} \\
\cmidrule(lr){2-16}
& \multicolumn{3}{c}{SUNTest}
& \multicolumn{3}{c}{NSGen}
& \multicolumn{3}{c}{\tool-C}
& \multicolumn{3}{c}{\tool-I}
& \multicolumn{3}{c}{\tool-S} \\
\cmidrule(lr){2-16}
VGG16
& 2092.2 & 0.65 & 0.66
& 4123.7 & 1.00 & 0.83
& \cellcolor{gray!40}4216.4 & 1.00 & \cellcolor{gray!40}0.41
& \cellcolor{gray!40}4389.2 & 1.00 & \cellcolor{gray!40}0.35
& \cellcolor{gray!40}\textbf{4752.0}
& {1.00}
& \cellcolor{gray!40}\textbf{0.31} \\

ResNet18
& 1981.5 & 0.59 & 0.86
& 4088.2 & 1.00 & 1.09
& \cellcolor{gray!40}4604.2 & 1.00 & \cellcolor{gray!40}0.39
& \cellcolor{gray!40}4845.6 & 1.00 & \cellcolor{gray!40}0.34
& \cellcolor{gray!40}\textbf{5264.6}
& {1.00}
& \cellcolor{gray!40}\textbf{0.32} \\

\midrule\midrule
& \multicolumn{15}{c}{\textbf{ImageNet}} \\
\cmidrule(lr){2-16}
& \multicolumn{3}{c}{ADAPT}
& \multicolumn{3}{c}{NSGen}
& \multicolumn{3}{c}{\tool-C}
& \multicolumn{3}{c}{\tool-I}
& \multicolumn{3}{c}{\tool-S} \\
\cmidrule(lr){2-16}
VGG19
& 3937.3 & 0.81 & 2.07
& 4633.4 & 1.00 & 2.16
& \cellcolor{gray!40}5206.5 & 1.00 & \cellcolor{gray!40}1.78
& \cellcolor{gray!40}\textbf{6149.7}
& {1.00}
& \cellcolor{gray!40}\textbf{1.51}
& \cellcolor{gray!40}5754.1 & 1.00 & \cellcolor{gray!40}1.77 \\

ResNet50
& 2842.5 & 0.83 & 2.21
& 4392.6 & 1.00 & 1.87
& \cellcolor{gray!40}5114.7 & 1.00 & 1.95
& \cellcolor{gray!40}\textbf{6235.7}
& {1.00}
& \cellcolor{gray!40}\textbf{1.77}
& \cellcolor{gray!40}5604.2 & 1.00 & 1.97 \\
\bottomrule
\end{tabular}
\caption*{\footnotesize \textit{Note:} Gray background highlights metrics where our methods outperform all baselines; Bold font indicates the best metrics.}
\vspace{-4mm}
\end{table*}

\begin{figure*}[t!]
  \centering
  \includegraphics[width=1\linewidth]{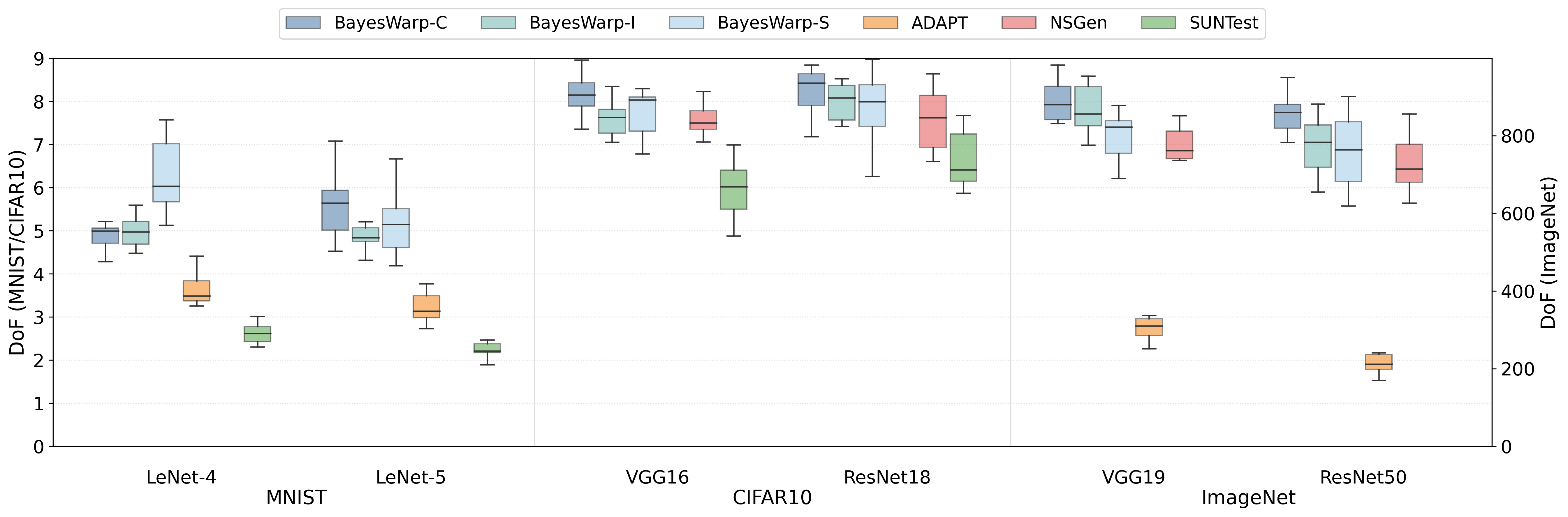}
  \vspace{-5mm}
  \caption{Comparison Results on DoF}
  \label{table:DoF-transposed-nodataset}
  \vspace{-5mm}
\end{figure*}

\noindent\textbf{\textit{Semantic Consistency Score (SCS)}}, following prior work~\cite{huang2024neuron}, measures the semantic similarity between an original input seed and its corresponding generated test case. 
Both images are encoded using the image encoder of a pretrained CLIP ViT-B/32 model~\cite{radford2021clip}, and SCS is computed as the cosine similarity between their image embedding vectors.

\noindent\textbf{\textit{Neuron Coverage (NC)}} measures the proportion of neurons activated during testing, following~\cite{pei2017deepxplore}.
A neuron is considered covered if its activation exceeds a predefined threshold for at least one generated test case.
NC is computed as the ratio of covered neurons to the total number of neurons, reflecting the overall breadth of internal-state exploration.

\noindent\textbf{\textit{Top-$k$ Neuron Coverage (TKNC)}} measures the percentage of unique neurons ranked among the top-$k$ activations per layer across all generated test cases, following~\cite{ma2018deepgauge, lee2020effective}, capturing how intensively different neurons are exercised.
We set $k=3$ for MNIST, $k=5$ for CIFAR-10, and $k=30$ for ImageNet, reflecting the increasing scale and complexity of the corresponding models: smaller networks require smaller $k$ to capture meaningful diversity, whereas larger networks demand higher $k$ to avoid underestimating coverage.

\noindent\textbf{\textit{Critical Neuron Coverage (CNC)}} measures the proportion of decision-critical neurons activated during testing, following~\cite{bai2024criticalfuzz}.
A neuron is deemed critical if its importance $\text{Imp}(n)=|\nabla_{a_n} s_c(x)\cdot a_n(x)|$ ranks within the top 10\% of its layer, where gradients are computed with respect to the softmax probability of the predicted class.
A critical neuron is counted as covered if its activation change exceeds a layer-wise threshold and its gradient magnitude lies above the 90th percentile.
CNC is computed as the ratio of covered critical neurons to all critical neurons.



\noindent\textbf{\textit {Accuracy (Acc)}}, as used in~\cite{pei2017deepxplore, wang2022bet}, measures whether generated test cases can improve prediction accuracy through fine-tuning.  

\section{Results and Analysis}

\subsection{RQ1: How effective is \tool\ in testing neural networks?}

We evaluate the effectiveness of \tool\ as a DNN testing technique from multiple perspectives.
In DNN testing, effectiveness is not only determined by the ability to trigger faults, but also by the diversity of exposed failures, the quality of generated test cases, and the extent to which decision-critical internal behaviors are exercised.
Accordingly, we assess \tool\ along four dimensions.
First, failure discovery capability and efficiency are measured using the Number of Failures (NoF), Failure-inducing Seed Rate (FSR), and Time per Failure (TPF), which capture how frequently, consistently, and efficiently failures are exposed under a fixed mutation budget.
Second, failure diversity is evaluated using the Diversity of Failures (DoF), indicating whether the method explores distinct erroneous behaviors rather than inducing similar failures.
Third, test case quality is assessed using Fréchet Inception Distance (FID) and Semantic Consistency Score (SCS), where FID measures distributional proximity to the original seed class and SCS quantifies semantic similarity between mutated inputs and their corresponding seeds.
Finally, as a white-box testing method, internal behavior exploration is examined using Neuron Coverage (NC), Top-k Neuron Coverage (TKNC) and Critical Neuron Coverage (CNC).
Together, these metrics provide a holistic evaluation of testing effectiveness. 
All experiments are repeated five times with different random seeds, and we report the mean results.

\begin{figure*}[t!]
  \centering
  \includegraphics[width=1\linewidth]{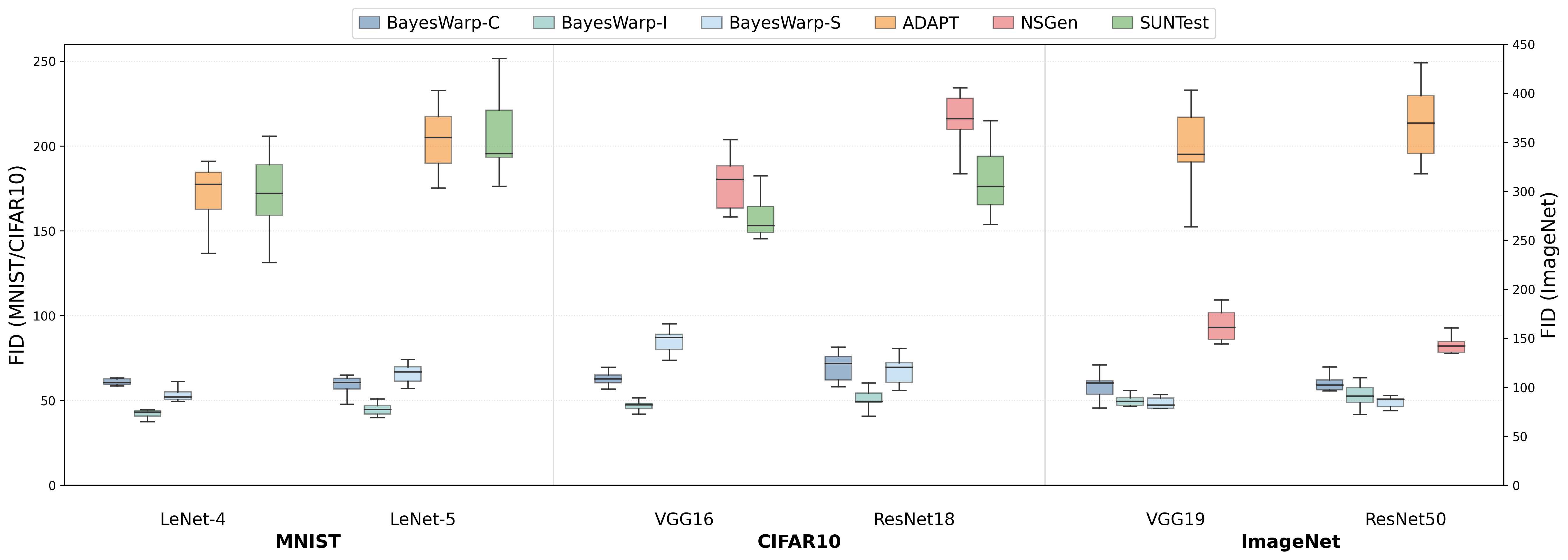}
  \vspace{-6mm}
  \caption{Comparison Results on FID}
  \vspace{-4mm}
  \label{table:FID}
\end{figure*}

\begin{figure*}[t!]
  \centering
  \includegraphics[width=1\linewidth]{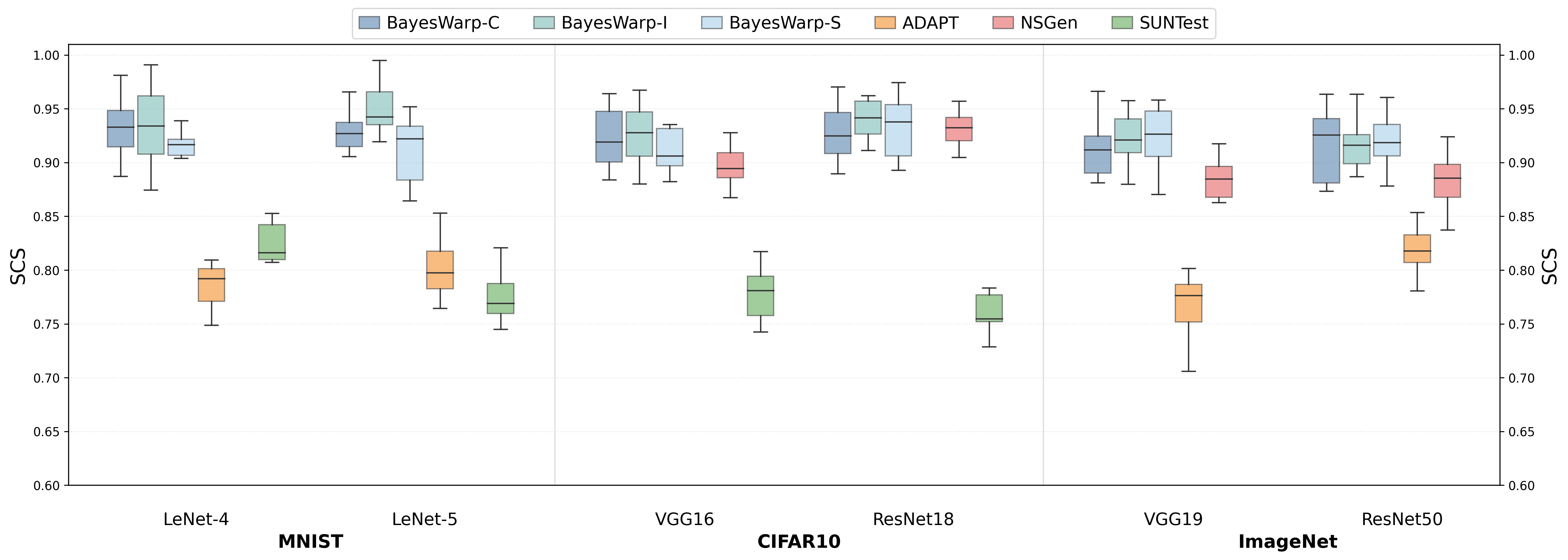}
  \vspace{-6mm}
  \caption{Comparison Results on SCS}
  \vspace{-4mm}
  \label{table:SCS}
\end{figure*}

\noindent\textbf{\textit{Capability and Efficiency.}}
Table~\ref{table:NoF-TPF-FSR} reports the comparison results on NoF, FSR, and TPF.
Overall, across all evaluated datasets and models, the three \tool\ variants achieve higher NoF than the baselines in most evaluated settings under the same mutation budget, indicating a stronger capability to expose failure-inducing behaviors. At the same time, baselines also demonstrate competitive performance in specific settings; for example, NSGen achieves relatively high NoF on CIFAR-10 and ImageNet, and ADAPT maintains stable performance across different models.
In addition, on MNIST, \tool\ consistently achieves higher FSR. On CIFAR-10 and ImageNet, all \tool\ variants reach the highest FSR, indicating that failures can be triggered for all selected input seeds. Baselines show performance differences across datasets and models: while NSGen also attains the highest FSR, ADAPT and SUNTest show lower success rates.
Both NoF and FSR are lower on MNIST than on CIFAR-10 and ImageNet. We do not interpret this as evidence that decision-critical localization is less useful on MNIST; rather, it reflects the more concentrated confidence structure and simpler local class competition of MNIST models. In such settings, each seed often admits fewer competitive alternative classes around the original prediction, which limits the number of distinct failure directions that can be effectively explored under the same mutation budget. By contrast, CIFAR-10 and ImageNet typically exhibit richer local competition among classes, making diversity-oriented exploration more productive.
Regarding efficiency, \tool\ generally requires less TPF than baselines while achieving higher NoF and FSR.
This suggests that the improved failure discovery is not achieved at the cost of increased computation, but rather through more effective use of the mutation budget.
In contrast, neuron-coverage-based methods require frequent queries of neuronal activation states during mutation, which incurs relatively higher runtime overhead. Although \tool\ incurs additional online costs (one-time saliency computation and lightweight surrogate updates), its decision-critical, uncertainty-aware search reduces wasted queries on irrelevant directions.

Notably, ADAPT and SUNTest guide mutations using neuron coverage signals, which facilitate internal activation exploration but do not distinguish whether mutated input regions are critical to the model’s final decision. As a result, a portion of their mutation budget is spent on activating neurons that have limited impact on prediction errors. NSGen explores diverse neuron activation patterns and demonstrates competitive performance on larger models. However, its mutation process does not explicitly focus on decision-critical input regions or adaptively guide mutations based on model confidence, which limits its effectiveness.
In contrast, \tool\ localizes decision-critical regions and adaptively guides mutations using model confidence through Bayesian optimization.
This enables \tool\ to focus exploration on input regions that are more likely to influence model decisions, which helps explain its favorable trade-off between failure discovery and efficiency in the evaluated settings.

\noindent\textbf{\textit{Diversity.}}
Figure~\ref{table:DoF-transposed-nodataset} reports the DoF across datasets and models.
On MNIST, all \tool\ variants achieve DoF values around 5-6 on LeNet-4 and LeNet-5, clearly higher than ADAPT and SUNTest, indicating that \tool\ can uncover more diverse failure behaviors even on simple models.
On CIFAR-10, \tool\ maintains DoF around 8--9 on VGG16 and ResNet18, consistently outperforming SUNTest and remaining competitive with, or better than, NSGen.
On ImageNet, the advantage becomes more pronounced: across VGG19 and ResNet50, \tool\ achieve higher DoF than ADAPT and NSGen, demonstrating good scalability to complex models and label spaces.

Overall, by focusing mutations on decision-critical regions and adaptively guiding exploration, \tool\ exposes a broader spectrum of failure modes under the evaluated settings.
In particular, adaptive target-class scheduling guides the search toward different competing classes, preventing concentration on a single dominant failure direction.


\noindent\textbf{\textit{Quality.}}
Figures~\ref{table:FID} and~\ref{table:SCS} report the quality of generated test cases in terms of FID and SCS, respectively.
Across all datasets and models, the three \tool\ variants generate test cases with lower FID than the baselines, indicating closer alignment with the original data distribution.
On MNIST and CIFAR-10, \tool\ maintain relatively lower FID, while baselines exhibit higher values.
On ImageNet, although FID increases for all methods due to dataset complexity, \tool\ variants consistently yield lower FID than ADAPT and NSGen on both VGG19 and ResNet50.
SCS results show a similar trend.
On MNIST, \tool\ variants achieve high semantic consistency, with median SCS values around 0.90.
These scores are consistently higher than those of ADAPT and SUNTest.
On CIFAR-10 and ImageNet, \tool\ maintains relatively higher SCS, whereas ADAPT shows a noticeable decline, indicating a higher likelihood of semantic deviation. In contrast, NSGen exhibits a more stable SCS trend, suggesting improved semantic consistency compared to ADAPT, but still below \tool.

Overall, these results indicate that \tool\ improves failure discovery without substantially degrading test case quality.
Notably, NSGen incorporates semantic-preserving mechanisms that help maintain semantic similarity.
However, its reliance on neuron-level activation guidance may cause mutations to deviate from the overall data distribution, leading to slightly higher FID.
By contrast, \tool\ constrains mutations within decision-critical regions and applies controlled mutations, preserving distributional proximity and semantic similarity while still inducing failures.

\begin{table*}[t!]
\centering
\footnotesize
\caption{Comparison Results on NC$\uparrow$, TKNC$\uparrow$ and CNC$\uparrow$.}
\label{table:Coverage}
\vspace{-1mm}
\setlength{\tabcolsep}{5pt}
\renewcommand{\arraystretch}{0.5}

\begin{tabular}{l | ccc ccc ccc ccc ccc}
\toprule
& \multicolumn{15}{c}{\textbf{MNIST}} \\
\cmidrule(lr){2-16}
& \multicolumn{3}{c}{ADAPT}
& \multicolumn{3}{c}{SUNTest}
& \multicolumn{3}{c}{\tool-C}
& \multicolumn{3}{c}{\tool-I}
& \multicolumn{3}{c}{\tool-S} \\
\cmidrule(lr){2-16}
\textbf{Model}
& NC & TKNC & CNC
& NC & TKNC & CNC
& NC & TKNC & CNC
& NC & TKNC & CNC
& NC & TKNC & CNC \\
\midrule
LeNet-4
& \textbf{74.6\%} & 25.6\% & 17.0\%
& 71.3\% & 22.8\% & 15.0\%
& 61.9\% & \cellcolor{gray!40}39.5\% & \cellcolor{gray!40}34.3\%
& 60.7\% & \cellcolor{gray!40}36.5\% & \cellcolor{gray!40}33.5\%
& 59.8\% & \cellcolor{gray!40}\textbf{39.8\%} & \cellcolor{gray!40}\textbf{37.4\%} \\
LeNet-5
& \textbf{73.8\%} & 25.0\% & 16.7\%
& 70.9\% & 23.7\% & 13.6\%
& 62.4\% & \cellcolor{gray!40}\textbf{39.3\%} & \cellcolor{gray!40}34.6\%
& 60.6\% & \cellcolor{gray!40}36.1\% & \cellcolor{gray!40}33.1\%
& 59.5\% & \cellcolor{gray!40}38.6\% & \cellcolor{gray!40}\textbf{36.4\%} \\
\midrule\midrule

& \multicolumn{15}{c}{\textbf{CIFAR-10}} \\
\cmidrule(lr){2-16}
& \multicolumn{3}{c}{SUNTest}
& \multicolumn{3}{c}{NSGen}
& \multicolumn{3}{c}{\tool-C}
& \multicolumn{3}{c}{\tool-I}
& \multicolumn{3}{c}{\tool-S} \\
\cmidrule(lr){2-16}
\textbf{Model}
& NC & TKNC & CNC
& NC & TKNC & CNC
& NC & TKNC & CNC
& NC & TKNC & CNC
& NC & TKNC & CNC \\
\midrule
VGG16
& 66.7\% & 18.2\% & 13.9\%
& \textbf{69.4\%} & 21.6\% & 16.1\%
& 56.8\% & \cellcolor{gray!40}\textbf{37.8\%} & \cellcolor{gray!40}33.6\%
& 55.9\% & \cellcolor{gray!40}37.6\% & \cellcolor{gray!40}\textbf{33.7\%}
& 54.6\% & \cellcolor{gray!40}36.1\% & \cellcolor{gray!40}33.1\% \\
ResNet18
& 64.1\% & 16.5\% & 12.1\%
& \textbf{67.8\%} & 19.5\% & 14.1\%
& 57.3\% & \cellcolor{gray!40}\textbf{38.8\%} & \cellcolor{gray!40}34.6\%
& 55.6\% & \cellcolor{gray!40}36.5\% & \cellcolor{gray!40}32.5\%
& 54.8\% & \cellcolor{gray!40}38.5\% & \cellcolor{gray!40}\textbf{34.8\%} \\
\midrule\midrule

& \multicolumn{15}{c}{\textbf{ImageNet}} \\
\cmidrule(lr){2-16}
& \multicolumn{3}{c}{ADAPT}
& \multicolumn{3}{c}{NSGen}
& \multicolumn{3}{c}{\tool-C}
& \multicolumn{3}{c}{\tool-I}
& \multicolumn{3}{c}{\tool-S} \\
\cmidrule(lr){2-16}
\textbf{Model}
& NC & TKNC & CNC
& NC & TKNC & CNC
& NC & TKNC & CNC
& NC & TKNC & CNC
& NC & TKNC & CNC \\
\midrule
VGG19
& 52.6\% & 25.8\% & 14.1\%
& \textbf{55.4\%} & 28.5\% & 16.5\%
& 44.8\% & \cellcolor{gray!40}\textbf{45.9\%} & \cellcolor{gray!40}\textbf{36.4\%}
& 43.7\% & \cellcolor{gray!40}45.8\% & \cellcolor{gray!40}34.5\%
& 41.9\% & \cellcolor{gray!40}42.7\% & \cellcolor{gray!40}34.5\% \\
ResNet50
& 37.9\% & 7.8\% & 6.7\%
& \textbf{40.8\%} & 10.0\% & 9.2\%
& 31.6\% & \cellcolor{gray!40}20.9\% & \cellcolor{gray!40}22.8\%
& 30.9\% & \cellcolor{gray!40}\textbf{22.7\%} & \cellcolor{gray!40}23.1\%
& 29.7\% & \cellcolor{gray!40}20.3\% & \cellcolor{gray!40}\textbf{23.7\%} \\
\bottomrule
\end{tabular}
\vspace{-4mm}
\end{table*}

\noindent\textbf{\textit{Coverage.}}
As shown in Table~\ref{table:Coverage}, although \tool\ is not designed to maximize conventional coverage metrics, it consistently attains higher TKNC and CNC in the evaluated settings. We note, however, that CNC is not entirely independent of our method design: both saliency-based localization and CNC involve gradient-informed notions of importance. Therefore, CNC should be interpreted as complementary evidence.
This pattern does not stem from maximizing activation counts, but is consistent with failure-oriented exploration that focuses on decision-critical input regions.
By prioritizing mutations within these regions, \tool\ generates a larger and more diverse set of failure-inducing test cases.
These failures are associated with different decision directions and error modes, which in turn tends to exercise multiple decision paths and engage a broader and more decision-relevant subset of neurons.
In contrast, coverage-driven methods primarily aim to activate new neurons without explicitly encouraging failure quantity or failure diversity.
As a result, increased activation breadth does not necessarily translate into richer failure behaviors or stronger engagement of decision-critical neurons.

Overall, these results indicate that prioritizing failure discovery and diversity is more effective for exercising decision-critical internal behaviors than directly optimizing coverage objectives.

\begin{figure*}[t!]
    \centering
    \includegraphics[width=1\textwidth]{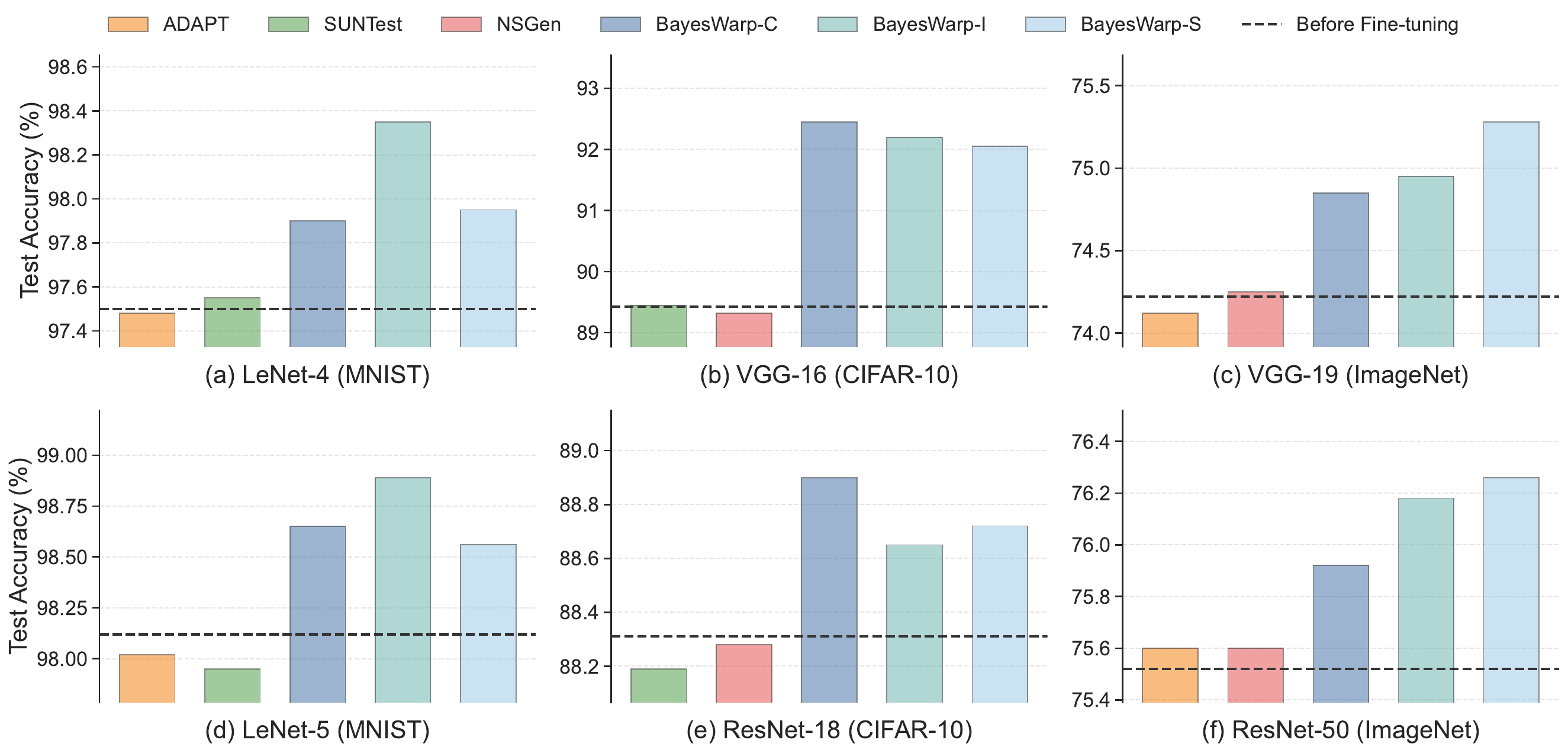}
    \vspace{-5mm}
    \caption{Test Accuracy Before and After Fine-Tuning for Different Testing Techniques.}
    \vspace{-2mm}
    \label{accuracy_comparison}
\end{figure*}

\subsection{RQ2: Can the generated test cases be leveraged to improve model performance?}

Following prior testing studies~\cite{wang2022bet, pei2017deepxplore}, we examine whether failure-inducing test cases can improve model performance through fine-tuning. For each method, we selected 1,000 failure-inducing test cases for fine-tuning. Failure cases were selected as evenly as possible across different input seeds and target classes, and supplemented when necessary to reach a fixed total of 1,000 cases. Models were fine-tuned for 10 epochs using the same optimizer and learning rate as in the original training.
As shown in Fig.~\ref{accuracy_comparison}, fine-tuning with test cases generated by \tool\ leads to larger test accuracy improvements than baselines on the test set.
Across all evaluated models, \tool\ yields clear accuracy gains.
These results suggest that the failures discovered by \tool\ provide more informative training signals than those generated by baseline testing techniques. Notably, all methods are fine-tuned using the same number of failure-inducing samples, isolating the effect of failure quality rather than data quantity.

These gains are unlikely to stem from data augmentation or coverage-driven input variation. Instead, \tool\ can expose model-intrinsic decision weaknesses by exploring decision-critical regions of the input space. Specifically, \tool\ uncovers a more diverse set of failure behaviors, as reflected by higher DoF, thereby reducing redundancy and revealing multiple distinct error modes. Meanwhile, the generated failures maintain low FID and high SCS, indicating that they remain distributionally and semantically close to the original data. Together, these properties suggest that the discovered failures correspond to meaningful decision-level deficiencies of the model rather than arbitrary or activation-driven mutations.
In contrast, coverage-guided methods primarily emphasize neuron activation, which can increase coverage without necessarily revealing decision-critical weaknesses, resulting in less informative failures for retraining. 

Overall, this analysis shows that \tool\ more effectively identifies intrinsic model weaknesses and enhances model performance.

\begin{table*}[t!]
\centering
\footnotesize
\caption{Ablation Study.}
\vspace{-1mm}
\label{table:ablation-full}
\renewcommand{\arraystretch}{0.6} 
\setlength{\tabcolsep}{9.6pt} 
\begin{tabular}{ll|rl|rl|rl|rl|rl}
\toprule
Model & Method & \multicolumn{2}{c}{NoF $\uparrow$} & \multicolumn{2}{c}{FSR $\uparrow$} & \multicolumn{2}{c}{DoF $\uparrow$} & \multicolumn{2}{c}{FID $\downarrow$} & \multicolumn{2}{c}{SCS $\uparrow$} \\
\midrule

\multirow{3}{*}{LeNet4}
& \textbf{\tool-S} & \textbf{3190.4} & {} & \textbf{0.47} & {} & \textbf{6.2} & {} & \textbf{52.8} & {} & \textbf{0.92} & {} \\
& w/o Localization & 1925.1 & \textcolor{red}{$\downarrow$1265.3} & 0.29 & \textcolor{red}{$\downarrow$0.18} & 4.9 & \textcolor{red}{$\downarrow$1.3} & 108.4 & \textcolor{red}{$\uparrow$55.6} & 0.79 & \textcolor{red}{$\downarrow$0.13} \\
& w/o Bayesian & 159.5 & \textcolor{red}{$\downarrow$3030.9} & 0.03 & \textcolor{red}{$\downarrow$0.44} & 0.8 & \textcolor{red}{$\downarrow$5.4} & 63.9 & \textcolor{red}{$\uparrow$11.1} & 0.89 & \textcolor{red}{$\downarrow$0.03} \\
\midrule

\multirow{3}{*}{LeNet5}
& \textbf{\tool-S} & \textbf{3828.9} & {} & \textbf{0.49} & {} & \textbf{5.2} & {} & \textbf{68.3} & {} & \textbf{0.92} & {} \\
& w/o Localization & 2088.4 & \textcolor{red}{$\downarrow$1740.5} & 0.26 & \textcolor{red}{$\downarrow$0.23} & 4.3 & \textcolor{red}{$\downarrow$0.9} & 145.2 & \textcolor{red}{$\uparrow$76.9} & 0.77 & \textcolor{red}{$\downarrow$0.15} \\
& w/o Bayesian & 191.4 & \textcolor{red}{$\downarrow$3637.5} & 0.04 & \textcolor{red}{$\downarrow$0.45} & 0.7 & \textcolor{red}{$\downarrow$4.5} & 76.5 & \textcolor{red}{$\uparrow$8.2} & 0.90 & \textcolor{red}{$\downarrow$0.02} \\
\midrule

\multirow{3}{*}{VGG16}
& \textbf{\tool-S} & \textbf{4752.0} & {} & \textbf{1.00} & {} & \textbf{8.1} & {} & \textbf{77.0} & {} & \textbf{0.92} & {} \\
& w/o Localization & 2580.6 & \textcolor{red}{$\downarrow$2171.4} & 0.54 & \textcolor{red}{$\downarrow$0.46} & 6.5 & \textcolor{red}{$\downarrow$1.6} & 172.4 & \textcolor{red}{$\uparrow$95.4} & 0.75 & \textcolor{red}{$\downarrow$0.17} \\
& w/o Bayesian & 237.6 & \textcolor{red}{$\downarrow$4514.4} & 0.05 & \textcolor{red}{$\downarrow$0.95} & 1.2 & \textcolor{red}{$\downarrow$6.9} & 89.6 & \textcolor{red}{$\uparrow$12.6} & 0.88 & \textcolor{red}{$\downarrow$0.04} \\
\midrule

\multirow{3}{*}{ResNet18}
& \textbf{\tool-S} & \textbf{5264.6} & {} & \textbf{1.00} & {} & \textbf{8.3} & {} & \textbf{67.3} & {} & \textbf{0.93} & {} \\
& w/o Localization & 3106.1 & \textcolor{red}{$\downarrow$2158.5} & 0.59 & \textcolor{red}{$\downarrow$0.41} & 7.4 & \textcolor{red}{$\downarrow$0.9} & 139.8 & \textcolor{red}{$\uparrow$72.5} & 0.79 & \textcolor{red}{$\downarrow$0.14} \\
& w/o Bayesian & 263.2 & \textcolor{red}{$\downarrow$5001.4} & 0.05 & \textcolor{red}{$\downarrow$0.95} & 1.0 & \textcolor{red}{$\downarrow$7.3} & 75.2 & \textcolor{red}{$\uparrow$7.9} & 0.91 & \textcolor{red}{$\downarrow$0.02} \\
\midrule

\multirow{3}{*}{VGG19}
& \textbf{\tool-S} & \textbf{5754.1} & {} & \textbf{1.00} & {} & \textbf{799.1} & {} & \textbf{80.9} & {} & \textbf{0.93} & {} \\
& w/o Localization & 2685.3 & \textcolor{red}{$\downarrow$3068.8} & 0.46 & \textcolor{red}{$\downarrow$0.54} & 214.2 & \textcolor{red}{$\downarrow$584.9} & 212.7 & \textcolor{red}{$\uparrow$131.8} & 0.73 & \textcolor{red}{$\downarrow$0.20} \\
& w/o Bayesian & 287.7 & \textcolor{red}{$\downarrow$5466.4} & 0.05 & \textcolor{red}{$\downarrow$0.95} & 32.1 & \textcolor{red}{$\downarrow$767.0} & 94.1 & \textcolor{red}{$\uparrow$13.2} & 0.90 & \textcolor{red}{$\downarrow$0.03} \\
\midrule

\multirow{3}{*}{ResNet50}
& \textbf{\tool-S} & \textbf{5604.2} & {} & \textbf{1.00} & {} & \textbf{763.3} & {} & \textbf{89.5} & {} & \textbf{0.92} & {} \\
& w/o Localization & 2914.2 & \textcolor{red}{$\downarrow$2690.0} & 0.52 & \textcolor{red}{$\downarrow$0.48} & 192.8 & \textcolor{red}{$\downarrow$570.5} & 199.5 & \textcolor{red}{$\uparrow$110.0} & 0.76 & \textcolor{red}{$\downarrow$0.16} \\
& w/o Bayesian & 280.2 & \textcolor{red}{$\downarrow$5324.0} & 0.05 & \textcolor{red}{$\downarrow$0.95} & 22.8 & \textcolor{red}{$\downarrow$740.5} & 99.3 & \textcolor{red}{$\uparrow$9.8} & 0.91 & \textcolor{red}{$\downarrow$0.01} \\

\bottomrule
\end{tabular}
\caption*{\footnotesize \textit{Note:} The red arrows indicate performance degradation.}
\vspace{-6mm}
\end{table*}

\subsection{RQ3: What is the impact of each main component in \tool?}

We construct two ablated variants of \tool-S:

\noindent \textbf{w/o Localization} removes the decision-critical region localization module: instead of restricting mutations to the localized critical region, it applies mutations over the whole input, while keeping the remaining pipeline unchanged.

\noindent \textbf{w/o Bayesian} removes Bayesian-guided search by replacing the acquisition-driven mutation selection with uniform random sampling in the mutation parameter space, while keeping critical region localization and the mutation query budget unchanged.

Table~\ref{table:ablation-full} shows that both components contribute to the effectiveness of \tool, but in different ways.
Removing localization leads to consistent degradation across all metrics. NoF decreases by 1,265-3,069 and FSR drops by 0.18-0.54, indicating that a relatively large portion of the mutation budget becomes ineffective when mutations are no longer constrained to decision-critical regions. DoF also declines, suggesting that the search drifts toward less informative directions and triggers fewer distinct failure modes.
At the same time, test case quality deteriorates: FID increases and SCS drops, showing that global mutations are more likely to deviate from the data distribution and break semantic consistency.
Replacing Bayesian-guided search with random mutation causes an even more severe drop in failure discovery.
Across all models, removing Bayesian-guided search leads to a substantial reduction in NoF, FSR, and DoF.
This indicates that, even with localization, random sampling is far less effective at refining mutations toward critical regions under the same budget. In contrast to localization removal, quality metrics are affected more mildly: FID increases slightly and SCS remains relatively high, suggesting that localization mainly provides distributional and semantic constraints, while Bayesian optimization mainly drives search efficiency and diversity.

Overall, localization contributes test case quality, whereas Bayesian-guided search determines how to explore within that region by improving number and diversity of failure.



\section{Discussion}

\begin{figure*}[t!]
    \centering
    \includegraphics[width=1\linewidth]{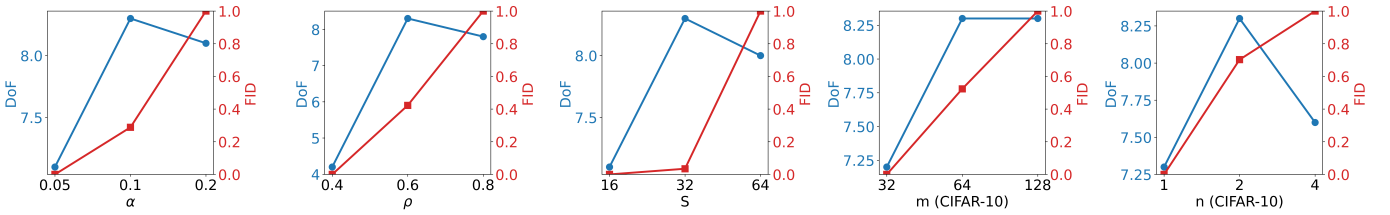}
    \vspace{-5mm}
    \caption{Impact of Key Hyperparameters.}
    \label{fig:hyper}
    \vspace{-3mm}
\end{figure*}

\begin{figure*}[t!]
  \centering
  \includegraphics[width=1\linewidth]{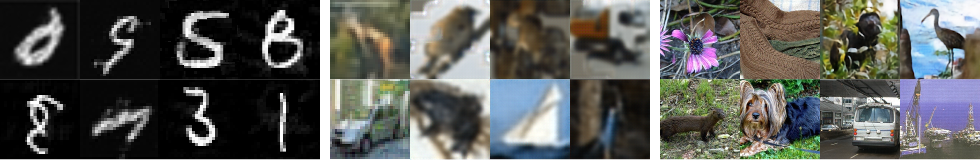}
  \vspace{-3mm}
  \caption{{Examples of Test Cases Generate by \tool}}
  \label{fig:example}
  \vspace{-3mm}
\end{figure*}

\begin{figure*}[t!]
    \centering
    \subfloat[LeNet-5~(MNIST)]{
        \includegraphics[width=0.325\textwidth]{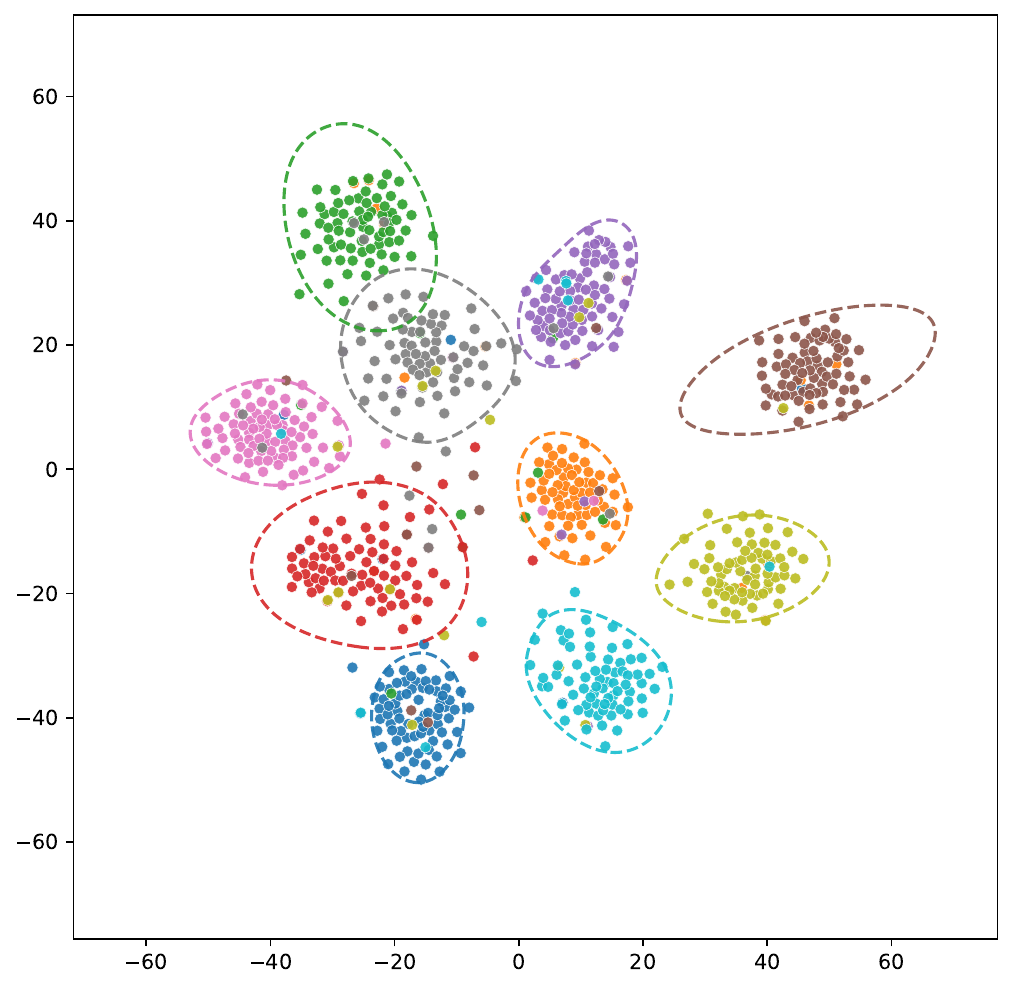}
    }
    \subfloat[ResNet-18~(CIFAR-10)]{
        \includegraphics[width=0.325\textwidth]{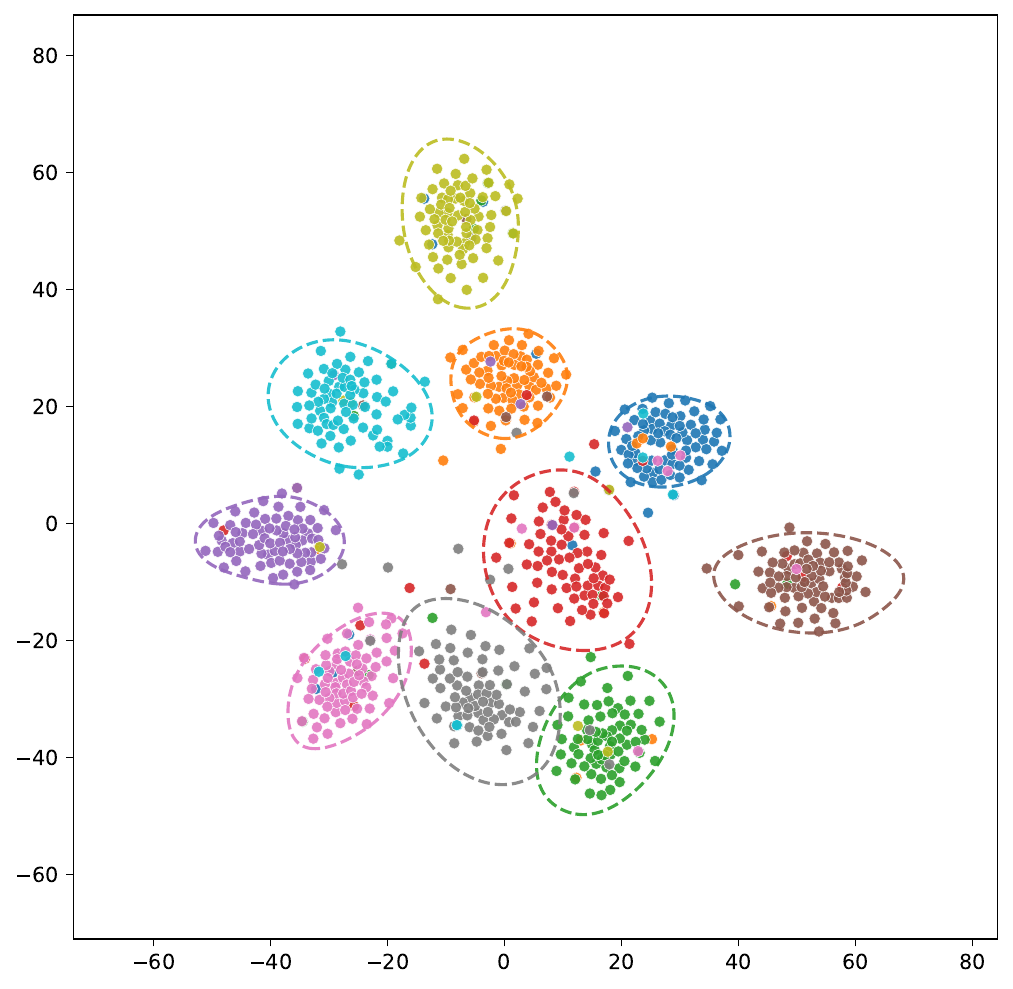}
    }
    \subfloat[ResNet-50~(ImageNet)]{
        \includegraphics[width=0.325\textwidth]{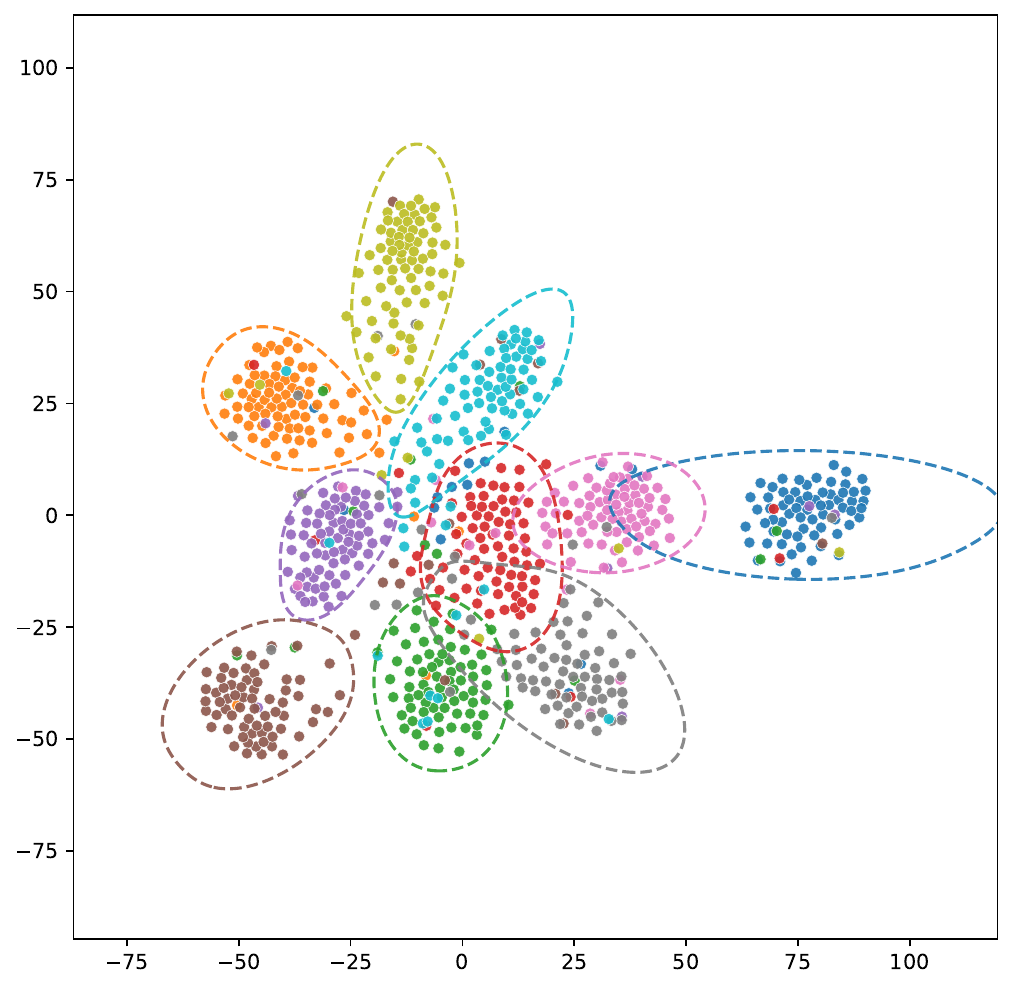}
    }
    \vspace{-1mm}
    \caption{t-SNE visualization of penultimate-layer feature representations.}
    \vspace{-3mm}
    \label{fig:tsne_discussion}
\end{figure*}

\noindent \textbf{Sensitivity Analysis of Key Hyperparameters.}
We evaluate the sensitivity of key hyperparameters in \tool-S.
For decision-critical region localization, we vary the saliency retention ratio $\alpha \in \{0.05, 0.1, 0.2\}$ and the maximum region coverage $\rho \in \{0.4, 0.6, 0.8\}$. 
For Bayesian-guided mutation, we vary the number of candidate mutations $S \in \{16, 32, 64\}$. 
For the SVGP surrogate, we vary $m \in \{32, 64, 128\}$ and $n \in \{1, 2, 4\}$. 
We center our sensitivity analysis on DoF and FID, which most directly reflect \tool’s core objectives.
As shown in Fig~\ref{fig:hyper}, we report the sensitivity analysis on CIFAR-10 (ResNet-18), which exhibit a clear behavior: DoF peaks near the default settings across all examined hyperparameters. In particular, varying $n$ and $\rho$ shows that coarser or more aggressive configurations do not yield DoF gains but instead increase FID. This shows that default configuration offers a balanced trade-off between diversity and fidelity, while alternative settings provide limited additional benefit.

\noindent \textbf{Case Study.}
Fig.~\ref{fig:example} shows failure test cases generated by \tool, randomly sampled from three datasets.
The generated test cases preserve the overall visual structure of the original inputs, while introducing localized mutations that are sufficient to induce failures.
On MNIST, digit shapes and stroke continuity are largely maintained.
On CIFAR-10 and ImageNet, object contours, textures, and scene context remain recognizable without obvious visual artifacts.

\begin{table}
\centering
\footnotesize
\vspace{-1mm}
\caption{Semantic stratification of failures.}
\vspace{-1mm}
\label{tab:scs_strat}
\setlength{\tabcolsep}{7.5pt}
\renewcommand{\arraystretch}{0.9}
\begin{tabular}{lcccc}
\toprule
SCS Range & ADAPT & SUNTest & NSGen & \tool-S \\
\midrule
High ($\geq$0.8) & 32.4\% & 38.1\% & 57.7\% & 71.3\% \\
Mid (0.6--0.8)   & 41.7\% & 36.4\% & 24.7\% & 21.4\% \\
Low ($<$0.6)     & 25.9\% & 25.5\% & 17.6\% & 7.3\% \\
\bottomrule
\end{tabular}
\vspace{-3mm}
\end{table}

\noindent \textbf{Semantic Similarity of Failures.}
While failures in \tool\ are detected using a prediction-inconsistency oracle, we conduct a post-hoc semantic analysis to assess whether they preserve semantic similarity. Fig.~\ref{fig:tsne_discussion} qualitatively visualizes penultimate-layer representations; it shows results of \tool\ on three models across MNIST, CIFAR-10, and ImageNet. Proximity in this space reflects representational similarity, and most generated test cases lie within or close to the original class clusters, indicating largely preserved semantics, with limited deviations expected from exploring decision-critical regions near class boundaries.
Complementing this view, Table~\ref{tab:scs_strat} reports a post-hoc SCS-based stratification of all failure-inducing test cases, averaged over all datasets and models under the same experimental settings as the preceding RQ1. Compared to baselines, \tool-S yields a higher proportion of high-consistency failures and fewer low-consistency ones, suggesting that most discovered failures preserve semantic similarity rather than resulting from excessive semantic drift.


\noindent \textbf{Impact of Different Localization Strategies.}
Although all three variants share the same Bayesian-guided testing pipeline, they differ in how decision-critical regions are localized, which leads to consistent but moderate performance differences.
\tool-C tends to identify coarse, class-discriminative regions at the feature-map level, which are effective for inducing failures but may miss fine-grained decision cues, resulting in slightly lower failure diversity and critical neuron coverage.
\tool-I provides finer-grained attributions along the input space, which helps preserve semantic consistency, but can introduce fragmented or noisy regions that reduce search stability under a fixed mutation budget.
In contrast, \tool-S aggregates saliency over multiple noisy samples, producing more stable and spatially coherent critical regions.
This stability allows the Bayesian optimizer to more reliably exploit decision-critical regions, leading to consistently higher NoF and DoF, as well as improved TKNC and CNC.
Overall, these results suggest that while \tool\ is robust to different localization strategies, stable and noise-robust critical-region estimation is particularly beneficial for efficient and diverse failure discovery.

\noindent \textbf{Interpretation of CNC.}
CNC is useful for examining whether generated test cases activate decision-relevant internal neurons, but it is not entirely independent from our testing pipeline. In particular, our localization stage uses saliency-based signals at the input level, whereas CNC is computed from neuron-level gradient-based importance. These two mechanisms are related, which may partially favor \tool\ on CNC. Nevertheless, they operate at different levels of abstraction: saliency determines where to perturb in the input space, while CNC evaluates which internal neurons are effectively exercised by the resulting test cases. We therefore interpret higher CNC not as standalone proof of superiority, but as supporting evidence consistent with our decision-critical testing objective.

\begin{table*}[t]
\footnotesize
\centering
\caption{Comparison between DNN Testing and Adversarial Attack}
\vspace{-2mm}
\label{tab:dnn_vs_adv}
\renewcommand{\arraystretch}{0.1}
\begin{tabular}{p{0.09\linewidth}p{0.41\linewidth} p{0.42\linewidth}}
\toprule
\textbf{Aspect} & \textbf{DNN Testing} & \textbf{Adversarial Attack} \\
\midrule
\textbf{Process} & Explores a wide range of inputs to examine whether a model behaves as expected, aiming to uncover diverse and meaningful failure. & Crafts targeted perturbations, often minimal or imperceptible, to induce misclassification under specified threat models. \\
\midrule
\textbf{Evaluation Metrics} & Failure diversity, state exploration, distributional and semantic proximity, and finetuning utility. & Attack success rate, perturbation magnitude, and perceptual imperceptibility. \\
\midrule
\textbf{Goal} & Reveal model weaknesses that generalize across inputs and support model improvement through failure-informed retraining. & Evaluate or strengthen robustness against specific adversarial threat models, focusing on boundary-level vulnerabilities.  \\
\bottomrule
\end{tabular}
\vspace{-1mm}
\end{table*}

\begin{table}
\centering
\footnotesize
\vspace{-2mm}
\caption{Comparison with Adversarial Attacks.}
\vspace{-1mm}
\label{tab:rq2_acc_dof_fid_autoattack}
\setlength{\tabcolsep}{4.8pt}
\renewcommand{\arraystretch}{0.5}
\begin{tabular}{l l l c c c}
\toprule
\textbf{Dataset} & \textbf{Model} & \textbf{Method} & \textbf{$\Delta$Acc (\%)} & \textbf{DoF $\uparrow$} & \textbf{FID $\downarrow$} \\
\midrule
\multirow{2}{*}{MNIST} 
& \multirow{2}{*}{LeNet5}
& \tool-S      & \textbf{+0.44}      & \textbf{5.2}  & \textbf{68.3} \\
& & AutoAttack   & \,-2.03             & 3.2           & 265.8 \\ 
\midrule
\multirow{2}{*}{CIFAR-10} 
& \multirow{2}{*}{ResNet18}
& \tool-S      & \textbf{+0.41}      & \textbf{8.3}  & \textbf{67.3} \\
& & AutoAttack   & \,-2.27             & 4.6           & 384.4 \\  
\midrule
\multirow{2}{*}{ImageNet} 
& \multirow{2}{*}{ResNet50}
& \tool-S      & \textbf{+0.74} & \textbf{763.3} & \textbf{89.5} \\
& & AutoAttack   & \,-1.64     & 317.2          &  472.9\\ 
\bottomrule
\end{tabular}
\vspace{-1mm}
\end{table}

\noindent \textbf{Comparison with Adversarial Attacks.}
Our work, in line with prior DNN testing methods~\cite{pei2017deepxplore, guo2018dlfuzz, lee2020effective, wang2022bet, dola2024cit4dnn}, is conceptually different from adversarial attacks~\cite{chakraborty2021survey, huang2017adversarial, villegas2024evaluating}. Adversarial attacks typically seek to induce misclassification with minimal perturbation under explicit threat models and are commonly evaluated using metrics such as attack success rate and perturbation magnitude. By contrast, \tool\ is designed as a testing technique: its goal is to uncover diverse, distribution-aligned failure modes and to improve model performance through failure-informed fine-tuning. Accordingly, \tool\ is specifically evaluated using DoF, TKNC, CNC, FID, and $\Delta$Acc, metrics that reflect failure diversity, internal-state exploration, data distributional proximity, and finetuning utility.
Table~\ref{tab:dnn_vs_adv} summarizes these conceptual differences. While adversarial attacks focus on boundary-adjacent perturbations, \tool\ explores a broader set of decision-critical regions, emphasizing the testing usefulness of generated failures rather than adversarial optimality.
To empirically illustrate these differences, we compare \tool-S with AutoAttack under matched budgets (10,000 generated samples per method). Our goal is not to outperform AutoAttack in adversarial criteria such as perturbation norms or success rates. Rather, we evaluate AutoAttack only under testing-oriented metrics to examine whether adversarially optimized perturbations translate into effective test inputs. We evaluate these using $\Delta$Acc, DoF, and FID, which directly correspond to our testing objectives: model-improvement utility, failure-mode diversity, and distributional proximity. As shown in Table~\ref{tab:rq2_acc_dof_fid_autoattack}, \tool\ consistently yields higher DoF, lower FID, and positive $\Delta$Acc after fine-tuning. AutoAttack, while effective as an adversarial attack, produces less diverse perturbations and degrades clean accuracy when its generated samples are used for retraining. These results highlight the distinction between testing-oriented mutations and adversarial attacks, which further confirms that adversarial optimization alone does not directly translate into effective DNN testing.

\section{Threats to Validity.}
We conducted a thorough review of our implementation to ensure correctness. For VGG and ResNet, we used PyTorch’s pre-trained weights and LeNet strictly follows its original design. All components were manually inspected to avoid code-level errors.

\noindent \textit{Internal validity.}
Validity may arise from randomness in mutation search and hyperparameter choices. We mitigate this by running each experiment with different random seeds and performing sensitivity analyses on key hyperparameters.
Our oracle relies on prediction inconsistency with respect to the original seed, following common practice in DNN testing. On large-scale datasets and high-confidence models, such an oracle may capture boundary-sensitive or semantically adjacent misclassifications rather than severe semantic errors. Accordingly, we use semantic similarity and feature-space visualizations only for post-hoc analysis, not for failure detection or oracle definition. While some identified failures involve only subtle semantic changes, they still reflect decision instabilities that are valuable for testing. For this reason, we additionally report FID, SCS, semantic stratification, and fine-tuning utility as complementary evidence, rather than relying on prediction inconsistency alone.

\noindent \textit{External validity.}
Our approach identifies decision-critical regions using saliency-based interpretability methods. Although we compare multiple saliency techniques and observe consistent performance trends, saliency maps and region merging criteria introduce heuristic factors that may affect region stability. Accordingly, the extracted regions should be interpreted as localized heuristic approximations. While we demonstrate robustness across multiple DNN architectures and saliency methods, the extraction behavior may differ for non-vision or non-convolutional models, and extending \tool\ to such domains remains future work. In addition, some evaluation metrics rely on pretrained models originally trained on natural images. When applied to MNIST, FID is used only as a relative indicator for comparing distributional consistency across methods, rather than as an absolute measure, following prior work~\cite{dola2024cit4dnn}. A further threat concerns baseline portability. Although we control the mutation budget, seed set, and oracle whenever possible, some prior white-box methods rely on architecture-specific or dataset-specific heuristics that limit fully uniform cross-setting reproduction. As a result, some comparisons in this study should be interpreted as reference comparisons.

\section{Related Work}

Recent neural network testing methods can be classified into white-box and black-box approaches based on model transparency. Coverage-based methods are typically used in white-box settings because they require internal information such as weights and parameters~\cite{pei2017deepxplore}, making them unsuitable for black-box scenarios. In contrast, black-box testing treats the model as a closed system and relies only on input-output behaviors, with existing methods guided by output feedback~\cite{odena2019tensorfuzz, xie2019deephunter}.

\textbf{\textit{Black-Box Testing Methods.}}
Black-box testing methods (e.g., random fuzzing or objective-driven mutation) operate purely on input--output behaviors and treat the model as a black box. Representative approaches include TensorFuzz~\cite{odena2019tensorfuzz}, DeepTest~\cite{pei2018deeptest}, DiffChaser~\cite{xie2019diffchaser}, DeepHunter~\cite{xie2019deephunter}, SINVAD~\cite{kang2020sinvad}, BET~\cite{wang2022bet}, ATOM~\cite{hu2023atom}, and CIT4DNN~\cite{dola2024cit4dnn}.
These methods mutate the {entire} input space, lacking reasoning about which regions drive the model's decision-making. Consequently, they require many queries, risk deviating from the natural data distribution, and limit coverage of diverse failure modes.
\textbf{\textit{White-Box Testing Methods.}}
White-box coverage-guided testing approaches such as DeepXplore~\cite{pei2017deepxplore}, DeepGauge~\cite{ma2018deepgauge}, DLFuzz~\cite{guo2018dlfuzz}, ADAPT~\cite{lee2020effective}, RobOT~\cite{wang2021robot}, NSGen~\cite{huang2024neuron} and SUNTest~\cite{guo2025white} mutate inputs according to neuron- or layer-level activation metrics. Such coverage signals can indeed reflect how extensively a model’s internalstates are exercised, providing a useful proxy for structural diversity during testing. However, coverage is not directly related to failure relevance: prior studies~\cite{harel2020neuron, yang2022revisiting} show that increasing activation coverage does not necessarily emphasize regions that critically influence the model’s predictions. Moreover, global mutation strategies commonly used in these methods may alter large portions of the input, risking distribution drift and reducing diversity.
\textbf{\textit{Saliency-Guided Learning.}}  
Beyond testing, saliency-guided adversarial methods leverage interpretability signals such as gradients or sensitivity maps to identify influential regions and craft perturbations that minimally alter the input~\cite{ismail2021improving, zhao2020scgan, wang2022adversarial}.
Their objectives differ from testing: they optimize for misclassification strength under norm constraints, not for failure diversity or coverage. Moreover, these attacks do not impose structural constraints on perturbed regions and exploration of decision-critical regions from a testing perspective.

\textbf{\textit{Summary.}}  
Overall, existing DNN testing methods suffer from two common limitations. First, they generally lack explicit decision-aware localization: black-box methods mutate the full input space, while white-box methods mainly rely on proxy signals such as neuron coverage, which are not always aligned with decision-critical regions. Second, they often struggle to balance efficiency and diversity under limited mutation budgets: random or global mutation wastes queries on irrelevant directions, whereas coverage-guided exploration may over-focus on dominant activation patterns and repeatedly uncover similar failures. 
Our work is motivated by these limitations. Rather than mutating the entire input or directly optimizing coverage signals, we formulate DNN testing as a process of exploring decision-critical regions in the model’s decision landscape. Based on this perspective, \tool\ combines saliency-based localization with uncertainty-aware Bayesian exploration, enabling more efficient and diversity-oriented failure discovery while maintaining distributional and semantic proximity to the original inputs.

\section{Conclusion}

In this work, we presented \tool, a Bayesian-guided white-box testing framework for neural networks that uncovers diverse model failures by focusing on decision-critical input regions. By combining structured saliency-based localization with uncertainty-aware optimization, \tool\ generates failure-inducing test cases that are both diverse and relatively well aligned with the original data distribution. More broadly, our results suggest the value of shifting from global or coverage-driven mutation toward decision-critical, diversity-oriented testing.

\bibliographystyle{IEEEtran}
\bibliography{main}

\end{document}